\documentclass[a4paper,fleqn]{cas-dc} 

\usepackage[numbers,sort&compress]{natbib} 

\usepackage{booktabs}
\usepackage{stfloats}
\usepackage{amssymb}
\usepackage{cleveref}
\usepackage{color}
\usepackage{caption}
\usepackage{algorithm}
\usepackage{algpseudocode}
\usepackage{amsmath}

\def\tsc#1{\csdef{#1}{\textsc{\lowercase{#1}}\xspace}}
\tsc{WGM}
\tsc{QE}

\usepackage{caption}
\begin{document}
	\let\WriteBookmarks\relax
	\let\printorcid\relax
	\def\floatpagepagefraction{1}
	\def\textpagefraction{.001}
	
	\shorttitle{}    
	
    \shortauthors{Yu et~al.}
	
	\title [mode = title]{JMT-YOLO: Multi-task Domain Adaptation via Segmentation-Guided Consistency Learning }

	\tnotemark[<tnote number>] 

	\author[1]{\textcolor[RGB]{0,0,1}{Zhenhua Yu}}
	\ead{zhenhuayu@xust.edu.cn}
	\author[1]{\textcolor[RGB]{0,0,1}{Jiahao Zhang}}
	\ead{23208223074@stu.xust.edu.cn}
	\author[1]{\textcolor[RGB]{0,0,1}{Wenchao Zhang}}
	\cormark[1]
	\ead{zhangwenchao@xust.edu.cn}
	\author[1]{\textcolor[RGB]{0,0,1}{Ou Ye}}
	\ead{oye0928@xust.edu.cn}
	\author[1]{\textcolor[RGB]{0,0,1}{Hengxiang Zhao}}
	\ead{24208207068@stu.xust.edu.cn}

	\address[1]{Institute of Systems Security and Control, College of Artificial Intelligence and Computer Science, Xi’an University of Science and Technology, Xi’an 710054, China}
        
	\cortext[1]{Corresponding author at: Institute of Systems Security and Control, College of Artificial Intelligence and Computer Science, Xi’an University of Science and Technology, Xi’an 710054, China.}

	\begin{abstract}
Domain adaptive object detection (DAOD) aims to reduce performance degradation due to distribution shifts between training (source) and testing (target) datasets. However, owing to insufficient modeling of fine-grained features and limited instance-level alignment, existing approaches become less effective under complex environmental conditions, such as extreme weather conditions and object occlusion. We overcome these challenges by proposing JMT-YOLO, a novel joint multi-task learning framework for DAOD. JMT-YOLO is based on the Mean Teacher architecture. It introduces an auxiliary semantic segmentation task and a masking mechanism to enhance the modeling of foreground–background semantic boundaries in the target domain. Specifically, fine-grained feature degradation is reduced by introducing a multiscale context-aware segmentation head equipped with a dense connection structure to effectively capture local and global contextual features. A mask-consistency loss is formulated by applying random masking operations to target images, thereby guiding the network to learn structural context and improving instance-level feature alignment. Extensive experiments on three benchmark domain adaptation tasks—Pascal VOC → Clipart1k, Cityscapes → Foggy Cityscapes, and KITTI → Cityscapes—demonstrate that JMT-YOLO achieves detection performances of 46.9\%, 57.3\%, and 61.0\%, respectively. These results demonstrate superior performance over existing state-of-the-art and baseline methods, thereby validating the effectiveness and generalization of JMT-YOLO. Our code will be available at \url{https://github.com/25043221/JMT-YOLO}.
	\end{abstract}
	
	
%
	\maketitle	
	
	\section{Introduction}
Deep learning–based object detection models have achieved remarkable success on standard benchmark datasets \cite{rf1,rf2,rf3,rf4}. These models generally assume that training (source) and testing (target) data are obtained from identical or closely related distributions. However, in practical scenarios, variations in environmental conditions (e.g., weather and illumination) and hardware settings (e.g., camera sensors and viewpoints) can result in considerable domain discrepancies between datasets \cite{rf5,rf6,rf7}. This phenomenon, referred to as domain shift, substantially limits the generalization capability of conventional object detectors when deployed in unseen or real-world environments.\par   
As increasing the diversity of training samples can alleviate domain shift to some extent, the collection and annotation of large-scale labeled data remain labor intensive and time-consuming \cite{rf8}. Therefore, domain adaptive object detection (DAOD), which reduces the reliance on target-domain annotations by transferring knowledge from a labeled source domain to an unlabeled or sparsely labeled target domain, has emerged as a promising solution \cite{rf9,rf10,rf11,rf12,rf58}.\par
DAOD techniques can be broadly categorized into two primary strategies: image-level and feature-level alignments. The former leverages style transfer or generative adversarial networks to reduce visual discrepancies between domains \cite{rf13,rf14,rf59}. By contrast, the latter aligns latent representations via adversarial training or distribution matching \cite{rf15,rf16}. As both the strategies have been effective in mitigating domain shift, they are insufficient in complex environments, particularly in scenarios involving occlusion, degraded visibility, or missing texture cues \cite{rf17}. Specifically, image-level methods prioritize global style consistency but often neglect the geometric and semantic variance of target objects, thereby losing fine-grained features and impairing the ability of the model to localize and recognize occluded or visually ambiguous objects \cite{rf52}. In addition, feature-level methods, although effective at global feature distribution matching, generally lack instance-level precision. In partially occluded or cluttered scenarios, these models can fail to construct holistic representations, resulting in unstable or inaccurate detections.\par
Herein, we address the aforementioned limitations by proposing JMT-YOLO, a novel DAOD framework grounded in joint multi-task learning. JMT-YOLO is built upon the Mean Teacher architecture and introduces two key components: (1) an auxiliary semantic segmentation task, designed to enrich pixel-level structural supervision and (2) a masking mechanism to enhance instance-level feature alignment by enforcing prediction robustness under partial occlusion. Specifically, a multiscale context-aware segmentation head is designed using a dense connection structure to capture rich local and global information, while a mask-consistency loss enables the model to learn semantically coherent representations under masked conditions. Extensive experiments on several challenging DAOD benchmark datasets demonstrate that JMT-YOLO consistently outperforms state-of-the-art approaches, thereby validating its effectiveness. The main contributions of this study are summarized as follows:\par
    (1) A joint multi-task DAOD framework is proposed, which integrates object detection and auxiliary segmentation tasks to enhance structural feature representation and domain alignment.\par
    (2) A masking mechanism is introduced to enhance robustness and instance-level alignment in complex environments via context-based structural learning.\par
    (3) Extensive experiments on three challenging DAOD datasets demonstrate that JMT-YOLO performs better in object detection tasks, which validates its effectiveness and generalization ability.\par

    The remainder of this paper is organized as follows: Section \ref{Related studies} reviews related studies. Section \ref{Proposed method} describes the proposed method. Section \ref{Experiments} presents and analyzes the experimental results. Section \ref{Conclusions} concludes the paper and outlines future directions.

	\section{Related studies}
	\label{Related studies}
This section reviews related studies from three key perspectives: DAOD, multi-task learning, and the Mean Teacher framework. These form the foundation of our proposed method. DAOD defines the problem setting, multi-task learning complements the use of auxiliary segmentation, and the Mean Teacher paradigm provides the core training strategy. Collectively, these threads contextualize the contributions of JMT-YOLO.
    
		\subsection{Domain adaptation object detection}
In recent years, DAOD has gained increasing attention owing to its potential to enhance the generalization performance of object detectors in cross-domain scenarios. By learning domain-invariant representations, DAOD alleviates the reliance on fully annotated target data and mitigates the impact of domain shift.\par
Early DAOD methods, such as domain-adversarial neural networks \cite{rf49}, introduce adversarial training mechanisms to align image-level feature distributions between source and target domains. Based on this idea, Chen et al. \cite{rf5} integrated instance-level domain discriminators and consistency regularization into Faster R-CNN \cite{rf2}, enabling the alignment of more granular cross-domains. Unlike global alignment strategies, instance-level discriminators assign domain labels to region proposals, thereby encouraging localized feature consistency. Moreover, image-level and instance-level consistencies regularization facilitate prediction stability across multiple granularities. Subsequent studies have primarily followed two-stage detection pipelines. For instance, Saito et al. \cite{rf6} proposed a strong-weak alignment strategy, in which strong alignment focuses on key samples, while weak alignment deals with the less prominent ones. He et al. \cite{rf8} developed multilevel adversarial training strategies to enhance the robustness of feature alignment. Guan et al. \cite{rf9} proposed uncertainty-aware adaptation methods by leveraging prediction confidence to selectively guide adaptation. Zhou et al. \cite{rf16} advanced this line of research by proposing multigranularity alignment across different representation levels.\par
Despite high accuracy of two-stage detectors, their computational complexity restricts their application in real-time settings. Therefore, recent studies are more focused toward single-stage detectors for DAOD. Hsu et al. \cite{rf18} introduced a center-aware alignment technique to improve instance localization in target domains. Chen et al. \cite{rf19} proposed I$^{3}$Net, which leverages implicit instance-invariant learning with multiscale pyramids and adversarial loss. Zhou et al. \cite{rf20} introduced SSDA-YOLO, a semi-supervised domain adaptation method based on YOLOv5 that incorporates knowledge distillation and consistency constraints. Guan et al. \cite{rf21} proposed a class-aware attention alignment module to enhance fine-grained feature adaptation.\par
Despite these advancements, most existing methods do not explicitly model semantic object boundaries. This limits their ability to distinguish fine-grained structures and align robustly at instance-level. The proposed method addresses these limitations by introducing an auxiliary semantic segmentation task and a masking mechanism for enhanced feature modeling and robust domain adaptation.

		\subsection{Multi-task learning}
    Multi-task learning (MTL) improves generalization by jointly training related tasks and enabling shared representations. In computer vision, MTL has been successfully applied to various tasks such as object detection, semantic segmentation, and pose estimation. Xu et al. \cite{rf22} proposed MTFormer, an MTL framework based on the Transformer architecture, which enables efficient information exchange via a cross-task attention mechanism, facilitating adaptive feature sharing. Fang et al. \cite{rf23} introduced ODFC-YOLO, which jointly trains dehazing and object detection tasks, considerably enhancing detection accuracy for unmanned aerial vehicle imagery in foggy environments. Zhu et al. \cite{rf24}  proposed a teacher-student MTL model that jointly learns object detection, drivable area segmentation, and lane line detection and thereby enhances performance in data-scarce environments. Liu et al. \cite{rf25} developed a saliency-guided MTL approach that integrates spatial saliency detection with temporal motion features for improving performance in infrared small object detection scenarios.\par

    Although MTL has been widely applied across a variety of tasks, Yang et al. \cite{rf55} further investigated it by integrating self-supervised and semi-supervised learning components into segmentation tasks. Their approach effectively addresses class imbalance in training data, aiming to enhance the performance of the model when labeled data are sparse. Similarly, Singh et al. \cite{rf56} proposed a multimodal MTL framework for effective computation, focused emotion detection, sentiment analysis, sarcasm recognition, and toxicity detection. Their method demonstrates the versatility of MTL in simultaneously tackling multiple related tasks and enhancing the performance of each task by leveraging shared representations.\par

    Herein, we integrate a semantic segmentation task as an auxiliary objective to provide pixel-level supervision. This enhancement enables our model to better capture structural and boundary information and ultimately improves the alignment in case of domain shifts.

    \begin{figure*}[hbtp]
    \centering
    \includegraphics[width=.9\textwidth]{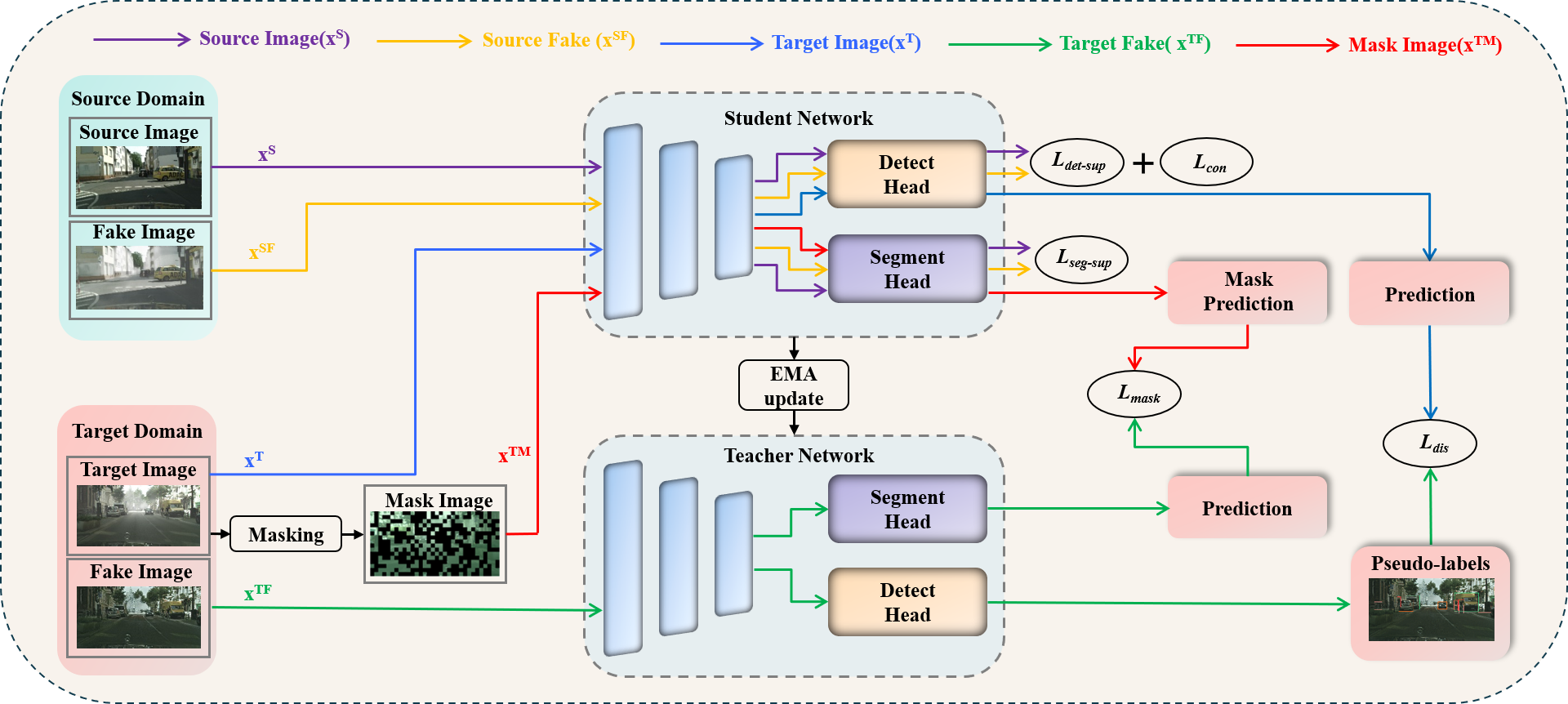}
    \caption{
    Overall architecture of the proposed JMT-YOLO framework. \textbf{Left}: pretrained contrastive unpaired translation (CUT) \cite{rf32} model is utilized to generate target-like images $x^{SF}$ from source images $x^{S}$ and source-like images $x^{TF}$ from target images $x^{T}$. Concurrently, a masked target image $x^{TM}$ is generated by applying a masking operation to $x^{T}$. \textbf{Middle}: a multi-task network, based on the MT framework, is constructed using the YOLOv5 \cite{rf38} detector. \textbf{Right}: corresponding loss functions are formulated based on predictions from multiple inputs (differentiated by color pathways) to support semi-supervised learning.}
    \label{fig:1}
    \vspace*{-0.8cm}
    \end{figure*}
    
        \subsection{Mean Teacher framework}
        The Mean Teacher (MT) framework, originally introduced by Tarvainen et al. \cite{rf26}, is a semi-supervised learning approach that maintains the teacher model using an exponential moving average (EMA) of the parameters of the student model. This EMA-based parameter smoothing enhances model prediction consistency and improves pseudo-label accuracy.\par

        Cai et al. \cite{rf27} extended the MT framework by introducing region-level, intra-graph, and inter-graph consistency to enforce spatial and structural coherence. Deng et al. \cite{rf28} proposed the unbiased MT (UMT) model, which combines style adaptation with confidence-based pseudo-label filtering to reduce noise from unreliable predictions. Li et al. \cite{rf29} developed the Adaptive Teacher model, which employs adversarial learning and differential augmentation (weak vs. strong) to enhance the pseudo-label quality. Kennerley et al. \cite{rf30} introduced 2PCNet for day-to-night adaptation, which applies a two-stage consistency strategy and specialized augmentation (NightAug) to address extreme illumination changes. Yang et al. \cite{rf31} proposed the Versatile Teacher model to address category imbalance and insufficient instance-level alignment. This framework uses category-aware confidence metrics to select reliable pseudo-labels and guide instance-level alignment using saliency maps.\par

        The MT framework has demonstrated considerable success in cross-domain object detection tasks by stabilizing pseudo-label learning and enforcing consistency constraints. The MT framework forms the theoretical foundation for the proposed method, which extends MT with segmentation-based supervision and a masking mechanism to enhance instance-level feature alignment under domain shift conditions.

	\section{Proposed method}
	\label{Proposed method}
        \subsection{Overall framework} 
    Given two distinct datasets in different domains: the source domain dataset, denoted as ${D_S} = \{ (x_i^S,y_i^S,m_i^S)\} _{i = 1}^{{n_S}}$, contains ${n_S}$ labeled samples. Each sample $x_i^S\in\mathbb{R}^{H\times W\times3}$ from the source domain \textit{S} is associated with an object detection label $y_i^S=\{(c_j^S,b_j^S)\}_{j=1}^{N_i}$ and a pixel-level segmentation mask $m_i^S \in {\{ 0,1\} ^{H \times W}}$. The object detection label $y_{i}^{S}$ comprises ${N_i}$ objects, where $c_{j}^{S}$ represents the class label and $b_j^S=(x,y,w,h)$ denotes the bounding box coordinates. The segmentation mask $m_{i}^{S}$ assigns a binary label to each pixel, where one as a value indicates foreground objects and zero indicates background regions. The target-domain dataset is denoted as ${D_T} = \{ (x_i^T)\} _{i = 1}^{{n_T}}$, where \textit{T} represents the target domain, $x_i^T\in\mathbb{R}^{H\times W\times3}$ represents the target-domain image, and ${n_T}$ indicates the number of target samples. As the target-domain data lack ground-truth (GT) annotations, a semi-supervised learning paradigm is adopted in our study. The domain adaptation approach aims to design a detector capable of transferring knowledge grasped from the source domain to the target domain, thereby achieving higher accuracy in the latter.\par

    A schematic of the proposed JMT-YOLO framework is illustrated in Fig. \ref{fig:1}. It comprises three main components: pseudo training images generation, consistency training, and joint optimization with a multi-task loss function. Specifically, the CUT model is first applied to generate target-like images $x^{SF}$ and source-like images $x^{TF}$. These enhance input diversity and bridge gaps at domain-level. In addition, random masks \cite{rf33} are applied to target images $x_{}^T$ to generate masked images $x_{}^{TM}$, which encourage the model to infer occluded content. Subsequently, all variants (original, translated, and masked images) are passed through a shared YOLOv5-based backbone and neck to extract multiscale features. These features are fed into two parallel branches: (1) a detection head for object classification and localization and (2) a segmentation head, which incorporates the proposed dense receptive field block (DenseRFB) to enhance semantic feature learning.\par

    Supervised learning is applied to \textit{S}, while pseudo-labels from the teacher model guide learning on \textit{T}. The student model is optimized via composite loss comprising detection, segmentation, and consistency losses, while the teacher model is updated using the EMA of student parameters.

        \subsection{Pseudo training images generation}

    In DAOD tasks, visual discrepancy often exists between the source and target domains. This domain shift causes the object detector trained on the source domain to perform poorly on the target domain, primarily owing to the misalignment in feature space, which makes it difficult for the detector to generalize unseen environments. Unlike other approaches, image-to-image translation methods alleviate the domain gap at the input level, which enables the detector to adapt to the characteristics of the target domain during training and thereby improve generalization \cite{rf53,rf54}. Inspired by Zhou et al. \cite{rf20}, we employed the CUT model to perform image style transfer at the input level and generate training samples that better match the target-domain distribution. Specifically, ${G_T}( \cdot )$ represent the image translation function from source to target domain and $G_{S}( \cdot )$ represent the reverse translation function from target to source domain. The construction of pseudo-images entails two directions: style transfer from the source domain to the target domain and vice versa. First, the CUT model is employed to translate source domain images $x^{S}$ to target-style images $x^{SF}$, as defined below:
\begin{equation}x^{SF}=G_T(x^S)
\end{equation}
where $x^{SF}$ retains the original annotations $(y_i^S,m_i^S)$ from the source domain, but its visual appearance is closer to the target domain. This enables the model to adapt to the target-domain distribution during training while allowing supervised learning. Similarly, source-style images $x^{TF}$ are generated by applying the inverse translator $G_{S}( \cdot )$ to the target-domain images $x^{T}$:
\begin{equation}x^{TF}=G_{S}(x^{T})
\end{equation}\par

\begin{figure} 
    \centering
    \includegraphics[width=\columnwidth]{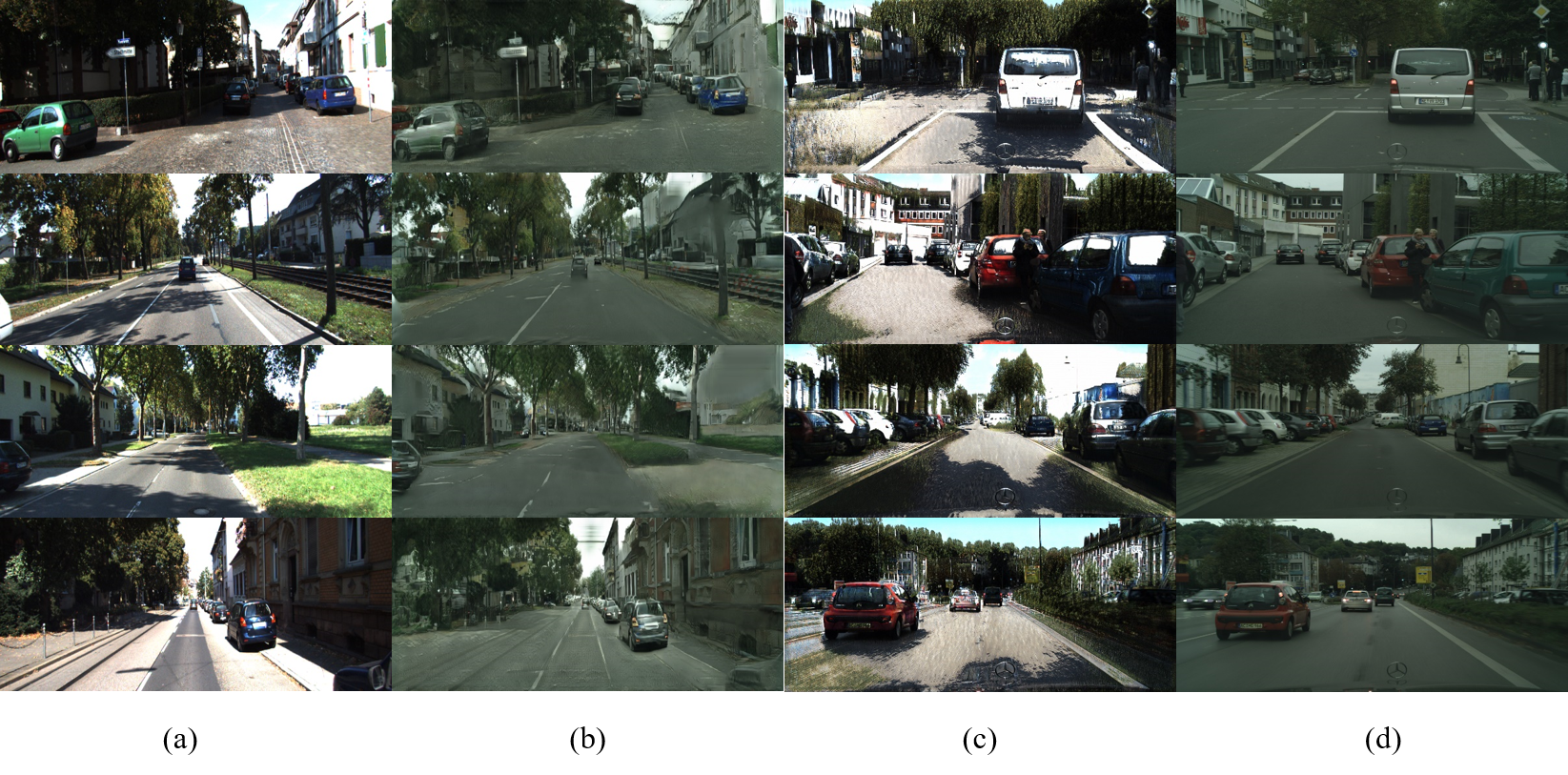}
    \caption{ Pseudo-images generated from KITTI and Cityscapes datasets. (a) Original images $x^{S}$ from the KITTI domain. (b) KITTI images translated into the target (Cityscapes) style images $x^{SF}$. (c) Cityscapes images translated into the source (KITTI) style images $x^{TF}$. (d) Original images $x^{T}$ from the Cityscapes domain.}
    \label{fig:2}
    \end{figure}

    Although $x^{TF}$ has a source-like appearance, it lacks GT annotations, as the original target samples are unlabeled. Therefore, it cannot be used directly for supervised training but is instead optimized using pseudo-labels generated by the teacher model, along with consistency constraints. As illustrated in Fig. \ref{fig:2}, the generated pseudo-images show considerable style changes while preserving structural information, thus enriching the training samples for subsequent multi-task learning.

    \subsection{Feature enhancement based on auxiliary segmentation}
     In cross-domain object detection tasks, insufficient feature representation often degrades model generalization performance. To address this issue, we propose a feature enhancement method that integrates an auxiliary segmentation task. Accordingly, a DenseRFB is designed to improve feature representation in cross-domain scenarios.\par

    The traditional receptive field block (RFB) \cite{rf50} can capture multiscale features using parallel convolutional branches. However, they exhibit clear limitations in cross-domain settings. First, the lack of sufficient interaction among branches leads to the inadequate fusion of local details and global context information. In addition, the absence of an effective hierarchical feature fusion mechanism hinders the ability of the model to capture important details as well as semantic information in cross-domain tasks. These limitations adversely affect the model performance under domain shifts.\par
    To overcome these problems, two key improvements are made to the traditional RFB module. First, a dense connection mechanism is introduced. The output of each layer is connected to the input of all subsequent layers. This design considerably enhances the interaction between multiscale features. In addition, it prevents the loss of feature information and makes the model more adaptable to complex domain shifts. Second, a hierarchical feature fusion strategy is adopted. This explicitly combines shallow detail features and deep semantic features, thereby improving the robustness of the model to domain discrepancies.\par
\begin{figure} 
    	\centering
    	\includegraphics[width=\linewidth]{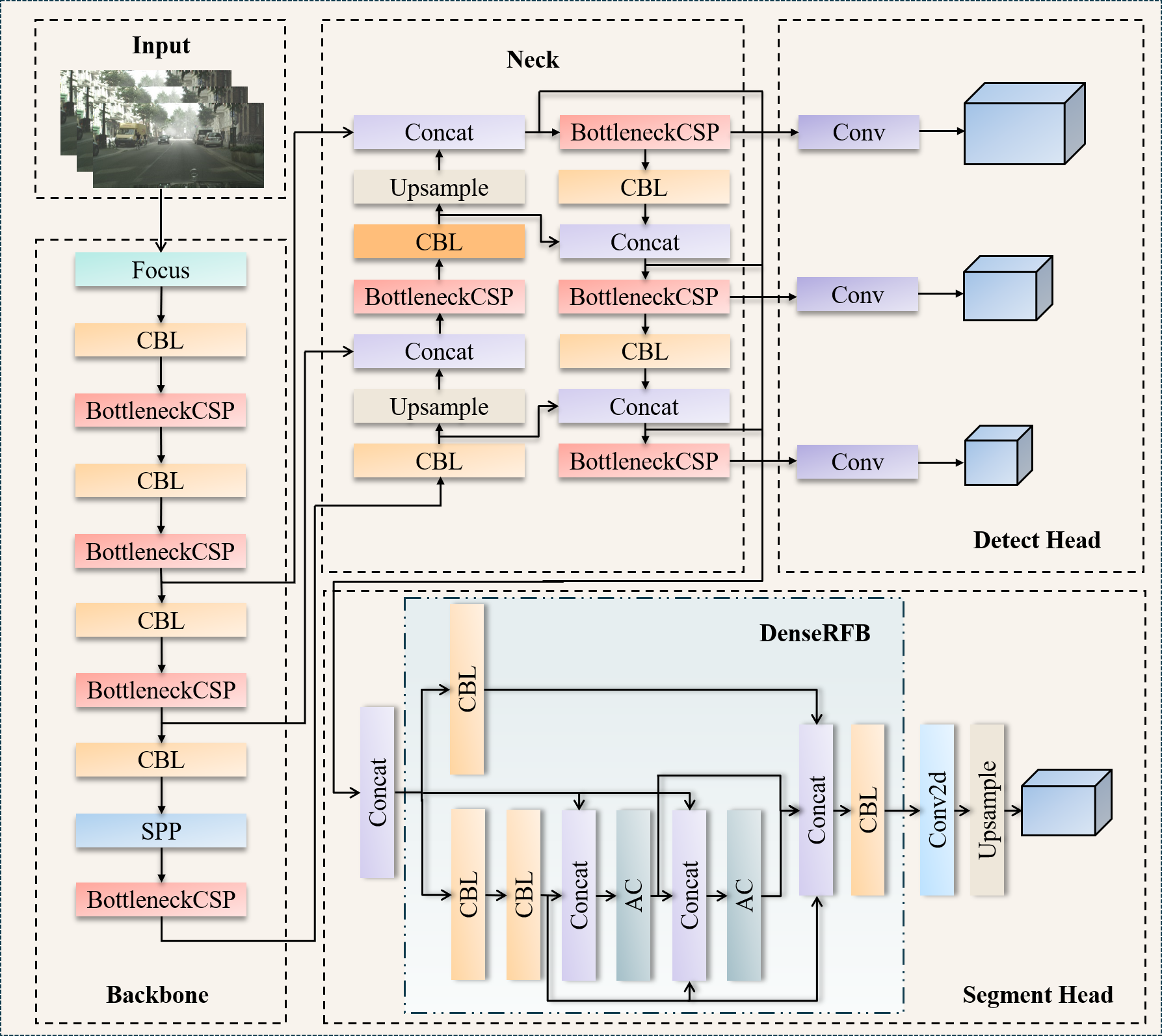}
    	\caption{Overview of the teacher-student network built on YOLOv5. A shared backbone and FPN–PAN neck feed two parallel heads: an unchanged detect head and an auxiliary Segment Head integrates the proposed DenseRFB for multiscale, context-aware features. Both the branches share the backbone/neck and are jointly optimized. “AC” denotes dilated convolution.}
    	\label{fig:3}
    \end{figure}

   \textbf{Backbone}: While implementing, the YOLOv5 architecture is adopted as the backbone for object detection. The backbone comprises a Focus module, multiple CBL modules, and BottleneckCSP modules (Fig. \ref{fig:3}). Once the input image passes through the backbone network, three feature maps are extracted at different scales. These are denoted as:
\begin{equation}\{\mathbf{F}_8,\mathbf{F}_{16},\mathbf{F}_{32}\}=\text{Backbone}(x)
\end{equation}
    where $\mathbf{F}_{k}\in\mathbb{R}^{\frac{H}{k}\times\frac{W}{k}\times C_{k}}$ represents the feature map with a downsampling factor of $k \in \{ 8,16,32\} $. \par
    \textbf{Neck}: The neck adopts an FPN + PAN structure to fuse features at different scales. Feature flow is achieved via upsampling, concatenation, and bottleneck modules to generate enhanced multiscale feature representations:
\begin{equation}\{\hat{\mathbf{F}}_8,\hat{\mathbf{F}}_{16},\hat{\mathbf{F}}_{32}\}=\mathrm{Neck}(\{\mathbf{F}_8,\mathbf{F}_{16},\mathbf{F}_{32}\})
\end{equation}\par

    \textbf{Detect head}: These enhanced features are fed into the detection head and auxiliary segmentation head for joint learning. For the detection branch, the original loss function of YOLOv5 is used. ${\hat x^S}$ denotes the predicted result of the student model for the input image  $x_{}^S$, and ${\hat x^{SF}}$ represents the predicted result of the student model for the input image $x_{}^{SF}$. The supervisory loss can be expressed as:
    \begin{equation} \mathcal{L}_{\mathrm{det-sup}}(\hat{x}^{S},b^{S},c^{S})=\mathcal{L}_{\mathrm{box}}(\hat{x}^{S};b^{S})+\mathcal{L}_{\mathrm{cls,obj}}(\hat{x}^{S};c^{S})
    \end{equation}
    \begin{equation}
    	\hat{\mathcal{L}}_{\mathrm{det-sup}}(\hat{x}^{SF},b^{S},c^{S})=\mathcal{L}_{\mathrm{box}}(\hat{x}^{SF};b^{S})+\mathcal{L}_{\mathrm{cls,obj}}(\hat{x}^{SF};c^{S})
    \end{equation}
    where ${\cal L}_{\det  - \sup }^{}$ refers to the detection supervision loss for the source domain images $x_{}^{S}$, $\hat {\cal L}_{\det  - \sup }^{}$ corresponds to the detection supervision loss for the target-style images $x_{}^{SF}$, $\mathcal{L}_{\mathrm{box}}$ denotes the GIoU loss for the predicted bounding boxes, and $\mathcal{L}_{\mathrm{cls,obj}}$ denotes the classification loss and focal loss for object scores, respectively, which are calculated using cross-entropy. Note that although the source domain images $x_{}^{S}$ and the target-style images $x_{}^{SF}$ have different data distributions at the scene level, they share the same label space. Inspired by Zhou et al. \cite{rf20}, we introduce a consistency loss $\mathcal{L}_{con}$ to constrain the outputs of the model within this label space. It is expressed as:
\begin{equation}
\mathcal{L}_{con}=\parallel\mathcal{L}_{\mathrm{det-sup}}(\hat{x}^{S},b^{S},c^{S})-\hat{\mathcal{L}}_{\mathrm{det-sup}}(\hat{x}^{SF},b^{SF},c^{SF})\|_{2}
\end{equation}\par

    \textbf{Segment head}: The segmentation branch is fundamentally designed to enhance instance-level alignment by leveraging pixel-wise semantic supervision. In cross-domain object detection, models frequently encounter domain-specific noise, such as artistic textures or environmental factors, which can blur crucial object structures and impede robust feature alignment. The proposed segmentation head addresses this problem by explicitly modeling domain-invariant structural cues, including object contours and foreground–background boundaries, using the novel DenseRFB module. The design of this module, which incorporates multibranch dense connectivity and dilated convolutions, enables capturing fine-grained local details and broader global context. This ensures that essential structural features are consistently preserved across different domains. By reframing instance alignment as the alignment of semantic pixel groups, the segmentation task directs the model to prioritize stable structural patterns over noisy variations. Consequently, this approach mitigates cross-domain feature misalignment and reinforces instance-level consistency, thereby functioning as a robust complement to the primary detection branch. Let the multiscale feature maps output by the neck be denoted as $\{\hat{\mathbf{F}}_{8},\hat{\mathbf{F}}_{16},\hat{\mathbf{F}}_{32}\}$. Following channel alignment and upsampling processing, they are concatenated to form a fused feature map $\hat{\mathbf{F}}_{f}$:
\begin{equation}
\hat{\mathbf{F}}_f=\left[\mathrm{Up}(\mathrm{Cv}_{1\times1}(\hat{\mathbf{F}}_{32}),\times4),\mathrm{Up}(\mathrm{Cv}_{1\times1}(\hat{\mathbf{F}}_{16}),\times2),\mathrm{Cv}_{1\times1}(\hat{\mathbf{F}}_8)\right]
\end{equation}
    where $\mathrm{Up}(\cdot,\times r)$ denotes the upsampling performed using bilinear interpolation, \textit{r} represents the upsampling factor, and $[ \cdot ]$ denotes the concatenation operation along the channel dimension.\par

    \textbf{DenseRFB}: The fused feature $\hat{\mathbf{F}}_{f}$ is the sole input to the DenseRFB module within the Segment Head. It first passes through a main branch, sequentially via 1 × 1 convolution and 3 × 3 standard convolution to obtain the initial features $\mathbf{F}_{b0}$:
\begin{equation}
\mathbf{F}_{b0}=\mathrm{Cv}_{3\times3}(\mathrm{Cv}_{1\times1}(\hat{\mathbf{F}}_f))
\end{equation}\par
    Based on this, two dilated convolution branches are constructed, using different dilation rates \textit{d}$_1$ and \textit{d}$_2$ (\textit{d}$_1$ = 2 and \textit{d}$_2$ = 3), to capture multiscale contextual information. Owing to the use of the dense connection strategy, the dilated convolution branches input the current feature map and integrate the outputs of all previous branches. The outputs of the second and third branches are denoted as:
\begin{equation}
\mathbf{F}_{b0}^{'}=\mathrm{Cv}_{1\times1}([\mathbf{F}_{b0},\mathrm{Cv}_{1\times1}(\hat{\mathbf{F}}_f)]),\mathbf{F}_{b1}=\mathrm{Cv}_{3\times3}^{d_1}(\mathbf{F}_{b0}^{'})
\end{equation}
\begin{equation}
\mathbf{F}_{b1}^{'}=\mathrm{Cv}_{1\times1}([\mathbf{F}_{b0},\mathbf{F}_{b1},\mathrm{Cv}_{1\times1}(\hat{\mathbf{F}}_f)]),\mathbf{F}_{b2}=\mathrm{Cv}_{3\times3}^{d_2}(\mathbf{F}_{b1}^{'})
\end{equation}
    where $\text{Cv}_{3\times3}^{d_i}(\cdot)$ denotes the 3 × 3 dilated convolution with dilation rate \textit{d}$_i$, and $\mathbf{F}_{b0}^{\prime}$,$\mathbf{F}_{b1}^{\prime}$ are the results of concatenating the outputs from multiple branches, which is followed by a 1 × 1 convolution for channel adjustment. \par

    The module further enhances information representation by introducing a direct connection branch that fuses the feature map $\hat{\mathbf{F}}_{f}$:
\begin{equation}
\mathbf{F}_{b3}=\mathrm{Cv}_{1\times1}(\hat{\mathbf{F}}_{f})
\end{equation}\par
    Finally, the outputs from all branches are fused along the channel dimension to generate the enhanced feature map $\mathbf{F}_{out}$:
\begin{equation}
\mathbf{F}_{out}=\mathrm{Cv}_{1\times1}([\mathbf{F}_{b0},\mathbf{F}_{b1},\mathbf{F}_{b2},\mathbf{F}_{b3}])
\end{equation}\par
    The final output of the DenseRFB module is used for the auxiliary semantic segmentation task, with the final output predictions $\hat{m}^{S}$ and $\hat{m}^{SF}$. The segmentation supervision loss is computed using the Dice loss function \cite{rf51}, which can be expressed as:
\begin{equation}
\mathcal{L}_{\mathrm{seg-sup}}(\hat{m}^{S},m^{S})=1-\frac{2\sum_{i}\hat{m}_{i}^{S}m_{i}^{S}}{\sum_{i}\hat{m}_{i}^{S}+\sum_{i}m_{i}^{S}+\epsilon}
\end{equation}
\begin{equation}
\hat{\mathcal{L}}_{\mathrm{seg-sup}}(\hat{m}^{SF},m^{S})=1-\frac{2\sum_{i}\hat{m}_{i}^{SF}m_{i}^{S}}{\sum_{i}\hat{m}_{i}^{SF}+\sum_{i}m_{i}^{S}+\epsilon}
\end{equation}
where ${\cal L}_{seg-sup}^{}$ represents the segmentation supervision loss for the source domain images  $x_{}^{S}$, $\hat{\mathcal{L}}_{\mathrm{seg-sup}}$ represents the segmentation supervision loss for the target-style images $x_{}^{SF}$, and $\epsilon$ is a small constant added for numerical stability.

        \subsection{Consistency training}
    In the MT architecture, the teacher and student models are instantiated from a source model (pretrained YOLOv5 architecture). During training, for each input image x in the source and target domains, we utilize the teacher model to obtain its classification scores $w_{cls}$, bounding box regression results $w_{box}$, and object confidence scores $w_{obj}$, which are later used as pseudo-labels to guide the training of the student model.\par
    For the optimization strategy, we use stochastic gradient descent and conduct supervised training of the student model using labeled samples from the source domain. Following MT setup, the parameters $\theta_{T}$  of the teacher model are updated using the EMA of the student model parameters $\theta_{S}$ as follows:
\begin{equation}
\theta_T=\gamma\theta_T+(1-\gamma)\theta_S
\end{equation} 
    where $\gamma$ is the decay factor. During pseudo-label generation, high-confidence bounding boxes are selected from the predictions of the teacher model as guidance signals for training on the target domains. This effectively reduces the prediction variance of the model in the target domain and enhances its robustness. For the detection branch, the teacher model receives an augmented target input $x_{}^{TF}$ and generates a prediction $\hat{x}^{TF}$. The student model receives a normal target input $x_{}^{T}$ and produces a prediction $\hat{x}^{T}$. A distillation loss is constructed based on the inconsistency between these two outputs:
\begin{equation}
\mathcal{L}_{dis}(\hat{x}^{T},\hat{x}^{TF})=\mathcal{L}_{\mathrm{det-sup}}(\hat{x}^{T},\mathcal{F}(\hat{x}^{TF}))
\end{equation} 
	where $\mathcal{F}(\cdot)$ denotes the teacher predictions $\hat{x}^{TF}$ after applying non-maximum suppression (NMS). During training, the teacher model is kept in evaluation mode. For each unlabeled target image, we first discard all predicted bounding boxes whose objectness confidence is lower than a fixed threshold $\delta$, and then perform class-aware NMS with an IoU threshold $\tau$. The remaining high-confidence boxes are used as hard pseudo-labels to supervise the student detector, enabling instance-level learning on the target domain.\par
	
	In the segmentation branch, we adopt a soft pseudo-labeling strategy to supervise the target domain without GT masks. Concretely, the teacher model produces pixel-wise class probability distributions for each target image. These probability maps are directly used as soft pseudo-labels: the student model is encouraged to match the teacher’s distribution at each pixel, which provides richer guidance than hard labels. However, naively relying on soft pseudo-labels has two potential drawbacks: (1) target-domain images may contain domain-specific noise that corrupts the teacher’s pixel-level predictions, and (2) the student can overfit to local pixel details without sufficiently leveraging global context. \par
	To alleviate these issues, a masking mechanism is introduced to improve robustness and promote context-aware learning. By applying random occlusions to local regions of target images, we compel the model to abandon its reliance on unreliable, domain-specific local pixels and instead infer occluded content through globally consistent contextual semantics. Since domain shift typically manifests as discrepancies in local textures while the global structural semantics of objects remain relatively invariant, this forced focus on structural dependencies naturally suppresses domain-specific variations. Consequently, the model is guided to extract domain-invariant features, which in turn enhances the critical instance-level alignment between the source and target domains. To explicitly enforce this semantic consistency under occlusion, a mask-consistency loss is defined between the predictions of the teacher and student models over the masked regions, formulated as follows:
\begin{equation}
\mathcal{L}_{mask}(\hat{x}^{TM},\hat{x}^{TF})=-\frac{1}{N}\sum_{i=1}^{N}\sum_{c=1}^{C}P_{i,c}^{T}\log P_{i,c}^{S}
\end{equation} 
    where $\hat{x}^{TM}$ and $\hat{x}^{TF}$ represent the prediction outputs of the student model for the masked image $x^{TM}$ and the teacher model for source-like image $x^{TF}$, respectively. \textit{N} indicates the total number of pixels in the image, and \textit{C}  indicates the total number of classes. $P_{i,c}^T$ represents the softmax probability predicted by the teacher model for class  \textit{c} at pixel \textit{i}, and  $P_{i,c}^S$ represents the corresponding softmax probability predicted by the student model. The complete training process is described in Algorithm~\ref{alg:jmt-yolo}.
    \begin{algorithm}[htbp]
    	\caption{Training process of JMT-YOLO}
    	\label{alg:jmt-yolo}
    	\textbf{Input:} Source domain data $D_S$, Target domain data $D_T$, Teacher model parameters $\theta_T$, Student model parameters $\theta_S$, EMA decay factor $\gamma$, Learning rate $\eta$, Number of training epochs $N$ \\
    	\textbf{Output:} Optimized teacher model parameters $\theta_T$
    	\begin{algorithmic}[1]
    		\State Initialize teacher model parameters: $\theta_T$
    		\For{epoch $t = 0$ to $N - 1$}
    		\For{each batch $(x^S, y^S,m^S),(x^{SF}, y^S,m^S) \in D_S$, $x^T,x^{TF} \in D_T$}
    		\State Generate pseudo-labels: $\mathcal{P}_T = \mathcal{F}(\hat{x}^{TF})$
    		\State Compute supervised detection loss $\mathcal{L}_{\text{det-sup}}$, $\hat{\mathcal{L}}_{\text{det-sup}}$ 
    		\State Compute segmentation loss $\mathcal{L}_{\text{seg-sup}}$, $\hat{\mathcal{L}}_{\text{seg-sup}}$
    		\State Compute distillation loss $\mathcal{L}_{\text{dis}}$
    		\State Compute mask consistency loss $\mathcal{L}_{\text{mask}}$
    		\State Compute total loss $\mathcal{L}_{\text{total}}$
    		\State Update student: $\theta_S = \theta_S - \eta \nabla \mathcal{L}_{\text{total}}$
    		\State Update teacher via EMA: $\theta_T = \gamma \theta_T + (1 - \gamma) \theta_S$
    		\EndFor
    		\EndFor
    		\State \Return $\theta_T$
    	\end{algorithmic}
    \end{algorithm}

          \subsection{Overall optimization}
    The design of the joint optimization function is divided into two parts: detection loss and segmentation loss, with different weight coefficients assigned to them. These weight coefficients are determined in the final ablation experiments. The design of the joint optimization loss is as follows:
\begin{equation}
\mathcal{L}_{total}=\lambda_{\det}\mathcal{L}_{\det}+\lambda_{seg}\mathcal{L}_{seg}
\end{equation} 
    where $\mathcal{L}_{\mathrm{det}}$  represents the detection loss and  $\mathcal{L}_{\mathrm{seg}}$ represents the segmentation loss, with $\lambda_{\mathrm{det}}$  and  $\lambda_{\mathrm{seg}}$ being their respective weights. Thus, $\mathcal{L}_{\mathrm{det}}$ includes the fully supervised detection losses $\mathcal{L}_{\mathrm{det-sup}}$ ,$\hat{\mathcal{L}}_{\mathrm{det-sup}}$ ,$\mathcal{L}_{\mathrm{con}}$ and $\mathcal{L}_{\mathrm{dis}}$ for the student model on source domain data and pseudo-source domain data. $\mathcal{L}_{\mathrm{seg}}$  includes the fully supervised segmentation losses $\mathcal{L}_{\mathrm{seg-sup}}$, $\hat{\mathcal{L}}_{\mathrm{seg-sup}}$  and $\mathcal{L}_{\mathrm{mask}}$ for the student model on source domain data and pseudo-source domain data.

	\section{Experiments}
	\label{Experiments}
    
    \subsection{Implementation details}
    All experiments were conducted on a PC running Ubuntu 20.04, which was equipped with a single NVIDIA RTX 3090 GPU (24 GB VRAM). The deep learning framework PyTorch 1.10.0 was used, with CUDA 11.1 and Python 3.8. YOLOv5-L was employed as the backbone detector. All input images were resized to 960 × 960 pixels. The network was trained using stochastic gradient descent with a momentum of 0.937 and a weight decay of $5\times10^{-4}$. The initial learning rate was set to $3\times10^{-3}$, and the batch size was two. The model was trained for 200 epochs under general settings, while models in the cross-camera adaptation configuration were trained for 50 epochs. For other hyperparameters, the design from the literature \cite{rf20} was followed: the IoU threshold was set to 0.3, the confidence threshold was set to 0.8, and the EMA smoothing coefficient was set to 0.99.
    \subsection{Datasets}
    We evaluated our approach across multiple benchmark datasets that covered natural and artistic domains as well as the following diverse driving scenarios:\par
    \textbf{Pascal VOC (2007 + 2012)} \cite{rf34}. Being the source domain for the real-to-artistic adaptation task, Pascal VOC provides 8,218 training and 8,333 validation images. The dataset included pixel-level segmentation annotations in addition to detection labels, and the segmentation task followed the same train and val partitions as those by detection. It covered 20 object categories shared with Clipart1k.\par
    \textbf{Clipart1k} \cite{rf13}. Clipart1k is an artistic-style dataset that is commonly used as the target domain for real-to-artistic adaptation. Following the official split, it contains 500 training and 500 validation images, each annotated with bounding boxes for the same 20 categories as those used in Pascal VOC.\par
   \textbf{Cityscapes} \cite{rf35}. Cityscapes is a large-scale urban driving dataset with high-quality detection and segmentation annotations. It provides 2,975 training and 500 validation images. The segmentation task used the same partitions as that of detection. Cityscapes was employed as the source domain for adverse-weather adaptation tasks and the target domain for cross-camera adaptation tasks.\par
    \textbf{Foggy Cityscapes} \cite{rf36}. Foggy Cityscapes was derived from Cityscapes by simulating foggy conditions. It retained the same partitions and annotation structure and was the target domain for adverse-weather adaptation tasks.\par
   \textbf{KITTI} \cite{rf37}. KITTI is a benchmark for autonomous driving with diverse camera viewpoints and higher-resolution imagery. In our experiments, we adopted 4,717 training and 520 validation images. As official segmentation labels were limited, we used a subset of 140 training and 60 validation images with pixel-level annotations to train the auxiliary segmentation branch.\par

    \subsection{Results and Analysis}
    Following previous studies \cite{rf5,rf20,rf31}, we evaluated the proposed method on three benchmark scenarios: (1) real-to-artistic adaptation, (2) adverse weather adaptation, and (3) cross-camera adaptation. All results were reported on the validation set of the target domain using mAP@0.5. In addition, we provided an Oracle result, where the model was trained and tested on the target domain, with training and validation splits strictly following the official dataset partitions.
    
    \begin{table*}[htbp]
    	\centering
    	\caption{The transfer results of PascalVOC → Clipart1k. The bold indicates the best performance, and underline indicates the second-best.}
    	\resizebox{\textwidth}{!}{
    		\begin{tabular}{l|c|cccccccccccccccccccccc}
    			\hline
    			\textbf{Method} & \textbf{Venue} & aero & bcycle & bird & boat & bottle & bus & car & cat & chair & cow & table & dog & hrs & bike & prsn & plnt & sheep & sofa & train & tv & all \\
    			\hline
    			SWDA~\cite{rf6} & CVPR19 & 26.2 & 48.5 & 32.6 & 33.7 & 38.5 & 54.3 & 37.1 & 18.6 & 34.8 & 58.3 & 17.0 & 12.5 & 33.8 & 65.5 & 61.6 & 52.0 & 9.3 & 24.9 & 54.1 & 49.1 & 38.1 \\
    			SCL~\cite{rf39} & arXiv19 & \textbf{44.7} & 50.0 & 33.6 & 27.4 & 42.2 & 55.6 & 38.3 & \underline{19.2} & 37.9 & 69.0 & 30.1 & 26.3 & 34.4 & 67.3 & 61.0 & 47.9 & 21.4 & 26.3 & 50.1 & 47.3 & 41.5 \\
    			DM~\cite{rf14} & CVPR19 & 25.8 & 63.2 & 24.5 & 42.4 & 47.9 & 43.1 & 37.5 & 9.1 & 47.0 & 46.7 & 26.8 & 24.9 & 48.1 & 78.7 & 63.0 & 45.0 & 21.3 & 36.1 & 52.3 & 53.4 & 41.8 \\
    			CRDA~\cite{rf40} & CVPR20 & 28.7 & 55.3 & 31.8 & 26.0 & 40.1 & 63.6 & 36.6 & 9.4 & 38.7 & 49.3 & 17.6 & 14.1 & 33.3 & 74.3 & 61.3 & 46.3 & 22.3 & 24.3 & 49.1 & 44.3 & 38.3 \\
    			HTCN~\cite{rf41} & CVPR20 & 33.6 & 58.9 & 34.0 & 23.4 & 45.6 & 57.0 & 39.8 & 12.0 & 19.7 & 51.3 & 21.1 & 20.1 & 39.1 & 72.8 & 63.0 & 43.1 & 19.3 & 30.1 & 50.2 & 51.8 & 40.3 \\
    			MEAA~\cite{rf42} & ACMMM20 & 31.3 & 53.5 & \textbf{38.0} & 17.8 & 38.5 & \underline{69.9} & 38.2 & \textbf{23.8} & 38.3 & 58.1 & 14.6 & 18.1 & 33.8 & \textbf{88.1} & 60.3 & 42.1 & 7.8 & 30.8 & \textbf{61.1} & \underline{58.7} & 41.1 \\
    			ATF~\cite{rf43} & ECCV20 & 41.9 & 67.0 & 27.4 & 36.4 & 41.0 & 48.5 & 42.0 & 13.1 & 39.2 & \underline{75.1} & 33.4 & 7.9 & 41.2 & 56.2 & 61.4 & 50.6 & \textbf{42.0} & 25.0 & 53.1 & 39.1 & 42.1 \\
    			I$^{3}$Net~\cite{rf19} & CVPR21 & 30.0 & 67.0 & 32.5 & 21.8 & 29.2 & 62.5 & 41.3 & 11.6 & 37.1 & 39.4 & 27.4 & 19.3 & 25.0 & 67.4 & 55.2 & 42.9 & 19.5 & 36.2 & 50.7 & 39.3 & 37.8 \\
    			PF-ATF~\cite{rf44} & TCSVT21 & 35.6 & 59.9 & 31.6 & 32.7 & 44.1 & 49.4 & 36.8 & 18.4 & 40.3 & \textbf{79.3} & 37.5 & 16.6 & \textbf{48.5} & 59.9 & 60.2 & 50.3 & \underline{35.6} & 23.2 & 49.4 & 46.9 & 42.8 \\
    			UMT~\cite{rf28} & CVPR21 & 39.6 & 59.1 & 32.4 & 35.0 & 45.1 & 61.9 & 48.4 & 7.5 & 46.0 & 67.6 & 21.4 & \textbf{29.5} & \underline{48.2} & 75.9 & \textbf{70.5} & 56.7 & 25.9 & 28.9 & 39.4 & 43.6 & 44.1 \\
    			TIA~\cite{rf45} & CVPR22 & 42.2 & 66.0 & \underline{36.9} & 37.3 & 43.7 & \textbf{71.8} & \underline{49.7} & 18.2 & 44.9 & 58.9 & 18.2 & \underline{29.1} & 40.7 & \underline{87.8} & 67.4 & 49.7 & 27.4 & 27.8 & \underline{57.1} & 50.6 & \underline{46.3} \\
    			DAVimNet~\cite{rf46} & arXiv25 & 33.2 & \underline{75.5} & 33.1 & 25.5 & 27.9 & \underline{69.9} & \textbf{50.1} & 16.9 & 40.8 & 47.0 & 32.6 & 24.1 & 32.5 & 77.0 & \underline{69.5} & 37.6 & 23.3 & \underline{41.2} & 57.0 & 48.2 & 43.8 \\
    			\hline
    			Source Only & - & 16.5 & 54.5 & 21.2 & 25.8 & 36.3 & 38.7 & 20.8 & 8.8 & 53.7 & 6.3 & 22.0 & 5.7 & 28.4 & 31.0 & 36.1 & 51.5 & 17.0 & 16.9 & 19.9 & 39.0 & 27.5 \\
    			SSDA-YOLO~\cite{rf20} & CVIU23 & 30.2 & \textbf{76.6} & 33.2 & \textbf{48.4} & \textbf{60.7} & 41.8 & 39.7 & 8.9 & \textbf{64.8} & 51.0 & \underline{38.7} & 25.5 & 36.8 & 44.5 & 59.9 & \underline{57.3} & 18.4 & \textbf{41.4} & 49.2 & 58.6 & 44.3 \\
    			\textbf{JMT-YOLO (Ours)} & - & \underline{43.9} & 65.4 & 23.3 & \underline{42.9} & \underline{59.9} & 57.0 & 48.2 & 7.3 & \underline{60.1} & 49.8 & \textbf{52.4} & 12.8 & 41.9 & 68.2 & 61.2 & \textbf{63.1} & 20.3 & 38.8 & 55.0 & \textbf{66.2} & \textbf{46.9} \\
    			Oracle & - & 21.7 & 74.3 & 34.6 & 39.8 & 47.1 & 74.8 & 57.8 & 22.3 & 47.4 & 52.7 & 34.1 & 39.1 & 29.8 & 58.7 & 75.3 & 46.2 & 45.4 & 27.0 & 48.0 & 41.0 & 45.8 \\
    			\hline
    		\end{tabular}
    	}
    	\label{tab:1}
    	
    \end{table*}
    \begin{figure*}[ht]
    	\centering
    	\includegraphics[width=\textwidth]{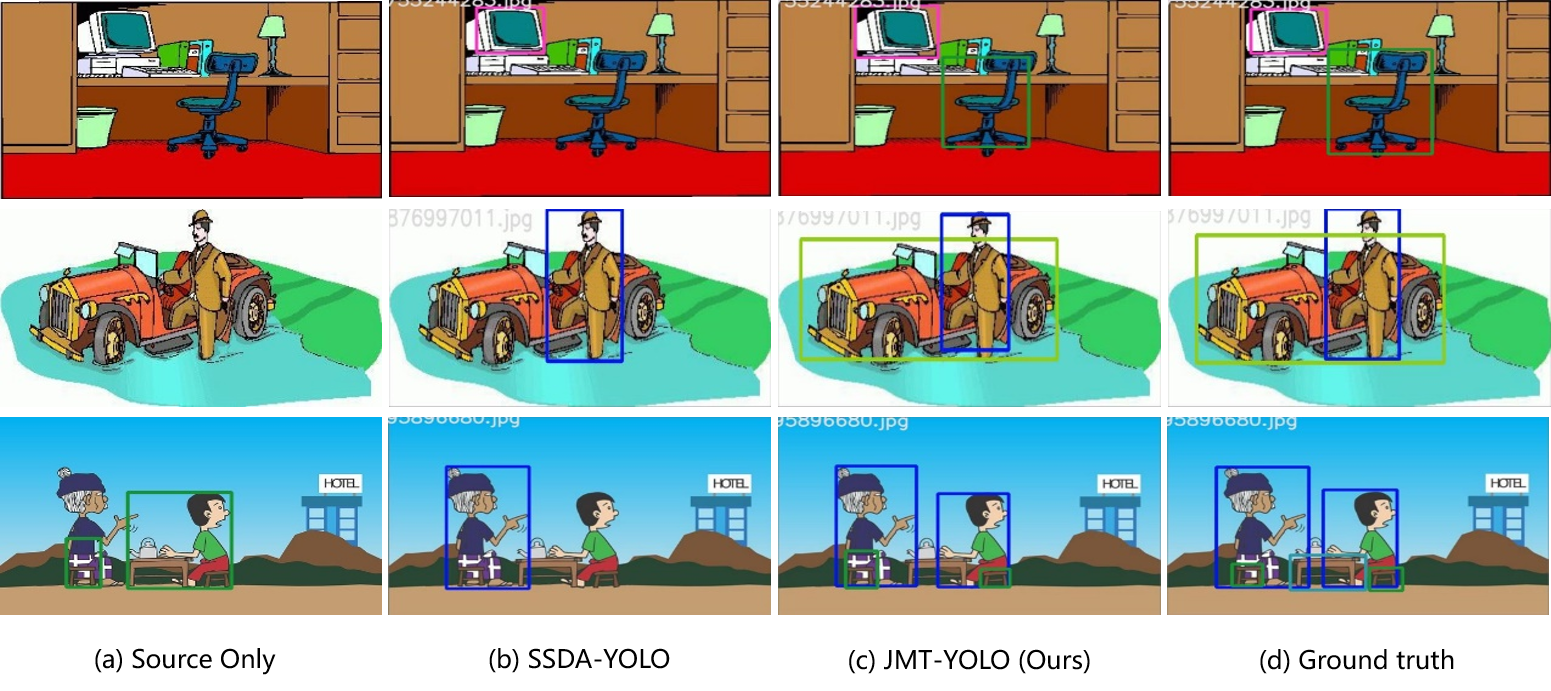}
    	\caption{Detection results of different methods on the Clipart1k dataset when PascalVOC is used as the source domain. JMT-YOLO localizes even better and accurately with fewer false positives.}
    	\label{picture/4.png}
    \end{figure*}

    \textbf{Real to artistic adaptation}. Artistic images substantially differ from real-life images in texture, color, and background, which sharply decreases the detection accuracy when a model trained on real data is directly applied to artistic domains. Table \ref{tab:1} presents the results of Pascal VOC → Clipart1k. The Source Only model demonstrates a sharp drop in performance (27.5\% mAP), indicating a large stylistic gap between real and artistic images. JMT-YOLO achieves 46.9\% mAP, surpassing SSDA-YOLO (44.3\%) and TIA (46.3\%). The improvement is primarily due to the auxiliary segmentation task, which provides spatial supervision for backbone features and enables the model to capture clearer object contours under artistic distortions. In addition, the DenseRFB module enriches multiscale representations and alleviates the loss of fine details due to style abstraction. As shown in Fig. \ref{picture/4.png}, JMT-YOLO produces clearer object boundaries and fewer false positives when compared with SSDA-YOLO. An improved appearance of texture-sparse categories, such as chair, person, and TV, confirms the effectiveness of segmentation-guided feature enhancement in handling style discrepancies.

\begin{figure*}[ht]
    \centering
    \includegraphics[width=\textwidth]{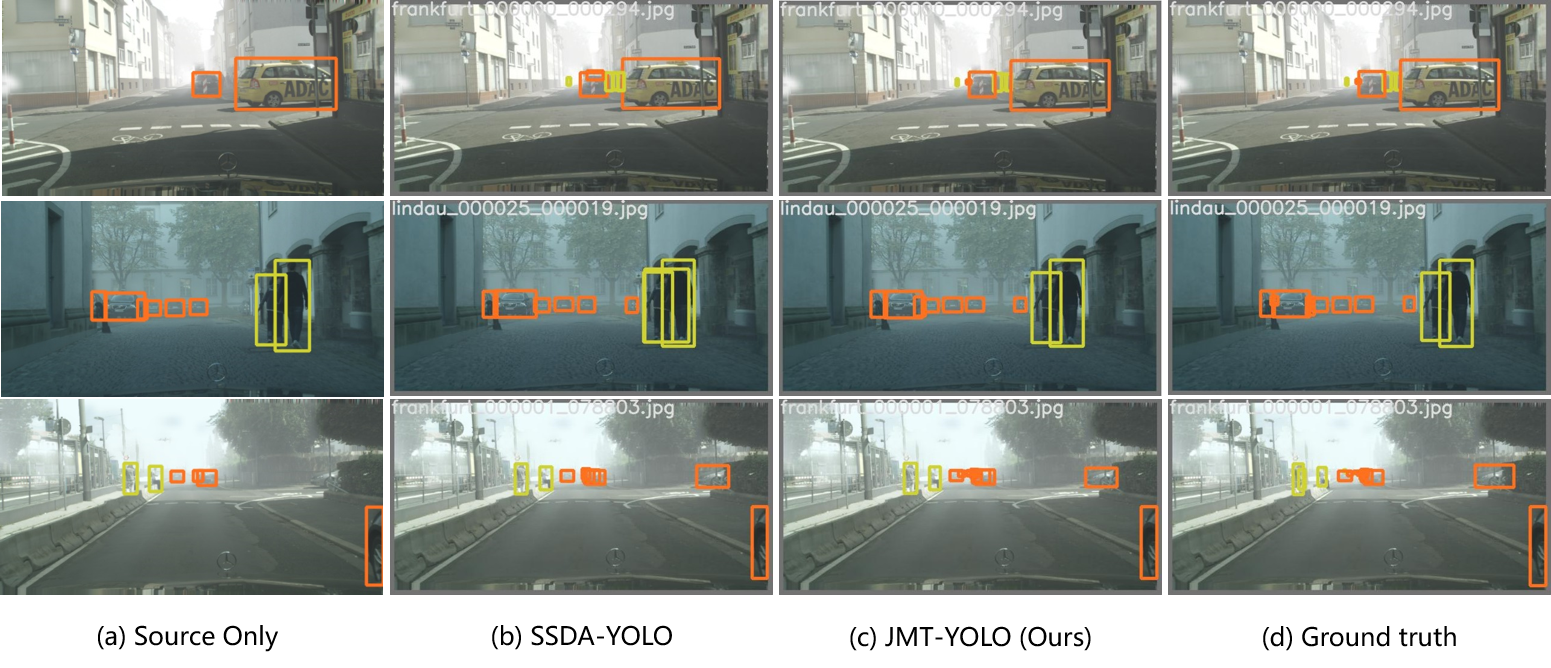}
    \caption{Detection results of different methods on Foggy Cityscapes dataset when Cityscapes is used as the source domain. JMT-YOLO shows improved robustness in detecting distant and occluded objects.}
    \label{picture/5.png}
\end{figure*}

     \textbf{Adverse weather adaptation}. Adverse weather conditions such as fog highly reduces image quality, causing a sharp decline in detection accuracy when models trained in clear weather are applied to degraded conditions. To evaluate the robustness of JMT-YOLO in such scenarios, we conducted experiments on Cityscapes → Foggy Cityscapes. Table \ref{tab:2} presents the corresponding results. The Source Only model reaches only 35.9\% mAP owing to visibility degradation due to fog. JMT-YOLO achieves 57.3\% mAP, outperforming SSDA-YOLO (55.9\%) and VT (51.1\%). The performance improvement primarily stems from the masking mechanism, which enforces consistency between detection and segmentation branches and helps the model to focus on object regions even under occlusion. Furthermore, the multi-task optimization enhances shared representations and reduces overfitting to clear weather features. As shown in Fig. \ref{picture/5.png}, JMT-YOLO detects distant and partially occluded objects more accurately. These results demonstrate that the proposed method performs robustly in adverse weather conditions.

\begin{table*}[ht]
\centering
\caption{The transfer results of Cityscapes → Foggy Cityscapes. The bold indicates the best performance, and underline indicates the second-best. FR-CNN refers to the use of the Faster R-CNN detector.}
\tabcolsep 0.4cm
\resizebox{\textwidth}{!}{
\begin{tabular}{l|c|c|cccccccc|c}
\hline
\textbf{Method} & \textbf{Venue} & \textbf{Detector} & \textbf{Bus} & \textbf{Bicycle} & \textbf{Car} & \textbf{Mcycle} & \textbf{Person} & \textbf{Rider} & \textbf{Train} & \textbf{Truck} & \textbf{All} \\
\hline
DA-Faster~\cite{rf5} & CVPR18 & FR-CNN & 35.3 & 27.1 & 40.5 & 20.0 & 25.0 & 31.0 & 20.2 & 22.1 & 27.6 \\
SWDA~\cite{rf6} & CVPR19 & FR-CNN & 36.2 & 35.3 & 43.5 & 30.0 & 29.9 & 42.3 & 32.6 & 24.5 & 34.3 \\
ATF~\cite{rf43} & ECCV20 & FR-CNN & 43.3 & 38.8 & 50.0 & 33.4 & 34.6 & 47.0 & 38.7 & 23.7 & 38.7 \\
UMT~\cite{rf28} & CVPR21 & FR-CNN & 56.5 & 37.3 & 48.6 & 30.4 & 33.0 & 46.7 & 46.8 & 34.1 & 41.7 \\
AT~\cite{rf29} & CVPR22 & FR-CNN & 58.4 & 53.0 & 63.8 & 34.4 & 43.2 & 52.7 & 34.5 & 32.4 & 46.6 \\
DSCA~\cite{rf17} & PR 2024 & FR-CNN & 52.6 & 35.2 & 68.8 & \textbf{52.7} & 53.9 & \underline{61.3} & 24.3 & 32.5 & 47.6 \\
VT~\cite{rf31} & PR 2025 & YOLOv5-L & \underline{63.0} & 44.1 & 65.9 & 38.7 & 49.2 & 51.1 & \textbf{56.3} & \underline{40.0} & 51.1 \\
\hline
Source Only & - & YOLOv5-L & 37.2 & 39.2 & 51.9 & 30.3 & 46.5 & 49.0 & 8.5 & 24.2 & 35.9 \\
SSDA-YOLO~\cite{rf20} & CVIU23 & YOLOv5-L & \underline{63.0} & \underline{53.6} & \underline{74.3} & \underline{47.4} & \underline{60.6} & \textbf{62.1} & 48.0 & 37.8 & \underline{55.9} \\
\textbf{JMT-YOLO (Ours)} & - & YOLOv5-L & \textbf{65.9} & \textbf{56.5} & \textbf{75.1} & 45.8 & \textbf{62.5} & 59.5 & \underline{52.3} & \textbf{40.7} & \textbf{57.3} \\
Oracle & - & YOLOv5-L & 61.2 & 52.1 & 77.2 & 47.8 & 62.8 & 60.3 & 50.3 & 46.2 & 57.2 \\
\hline
\end{tabular}
}
\label{tab:2}
\end{table*}

\begin{figure*}[ht]
    \centering
    \includegraphics[width=\textwidth]{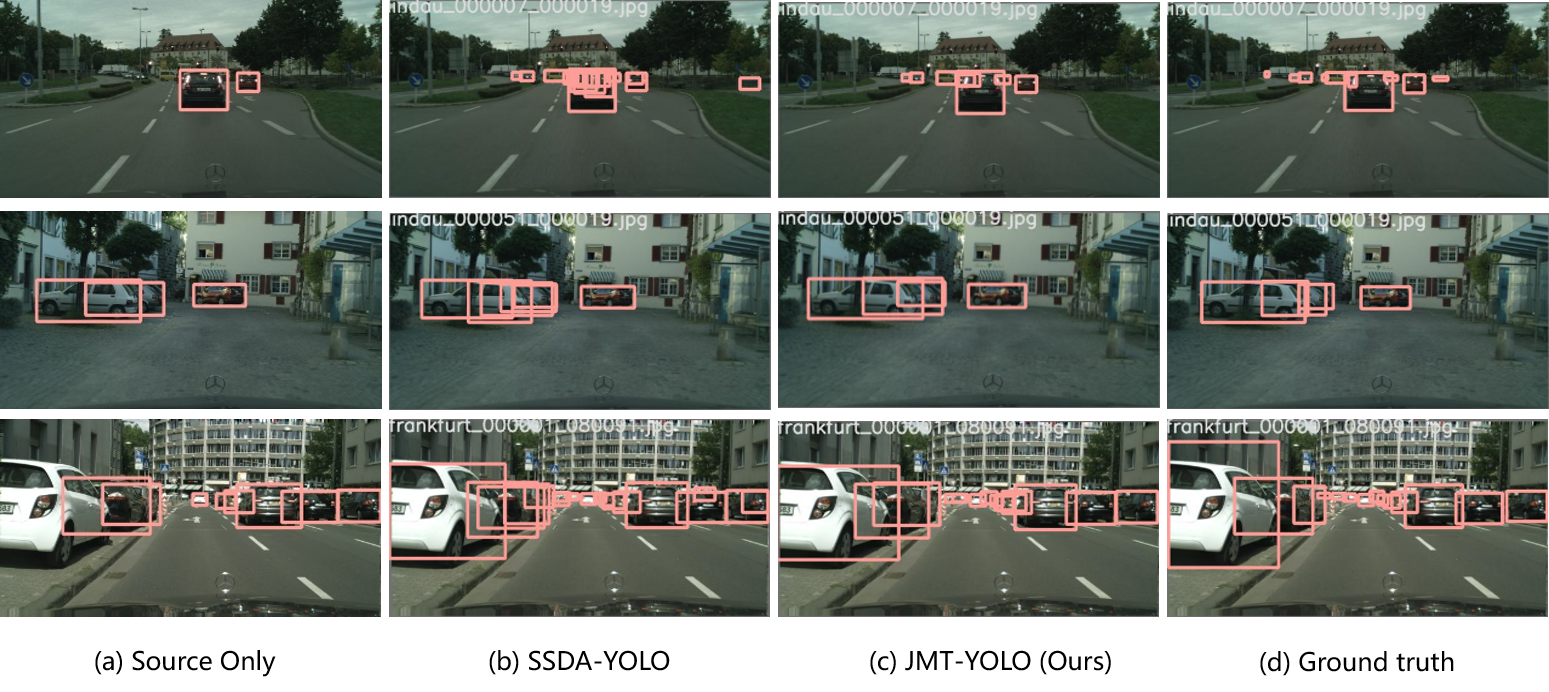}
    \caption{Detection results of different methods on Cityscapes dataset when KITTI is used as the source domain. Our method JMT-YOLO achieves more precise localization and fewer false detection.}
    \label{picture/6.png}
\end{figure*}

     \textbf{Cross-camera adaptation}. In cross-camera scenarios, differences in viewpoints, resolutions, and lighting conditions often result in a considerable drop in detection accuracy when a model trained on one camera is applied to another. To evaluate the adaptability of JMT-YOLO under such domain shifts, we conducted experiments on KITTI → Cityscapes (Table \ref{tab:3}). The Source Only model achieves 58.3\% mAP on the car category, while JMT-YOLO improves the performance to 61.0\%, outperforming most recent approaches and approaching the performance level of VT (61.9\%). It is worth noting that this result is obtained with limited segmentation annotations (only 200 images with GT masks). As demonstrated in our ablation study (Section~\ref{sec:sam} and \ref{sec:sam_noisy} ), leveraging the Segment Anything Model (SAM) \cite{rf57} to generate full-dataset pseudo masks with GT-box prompts effectively mitigates this constraint, further boosting the performance to 62.0\%. As shown in Fig. \ref{picture/6.png}, JMT-YOLO reduces false detections and improves localization precision when compared with SSDA-YOLO, confirming the advantage of segmentation-guided consistency. 

\begin{table}[ht]
\centering
\caption{The transfer results of KITTI → Cityscapes. The bold indicates the best performance, and underline indicates the second-best.}
\scriptsize
\setlength{\tabcolsep}{8pt}
\begin{tabular}{l|c|c|c}
\hline
\textbf{Method} & \textbf{Venue} & \textbf{Detector} & \textbf{Car AP} \\
\hline
EPM~\cite{rf18} & ECCV20 & YOLOv5-L & 58.1 \\
DAYOLO~\cite{rf12} & PMLR21 & YOLOv5-L & 57.4 \\
MSDAYOLO~\cite{rf47} & ICIP21 & YOLOv5-L & 56.5 \\
AT~\cite{rf29} & CVPR22 & YOLOv5-L & 59.0 \\
CMT~\cite{rf48} & CVPR23 & YOLOv5-L & 53.2 \\
VT~\cite{rf31} & PR25 & YOLOv5-L & \textbf{61.9} \\
\hline
Source Only & - & YOLOv5-L & 58.3 \\
SSDA-YOLO~\cite{rf20} & CVIU23 & YOLOv5-L & 59.6 \\
\textbf{JMT-YOLO (Ours)} & - & YOLOv5-L & \underline{61.0} \\
Oracle & - & YOLOv5-L & 72.2 \\
\hline
\end{tabular}
\label{tab:3}
\end{table}

    \subsection{Ablation studies}
    All ablation analyses were performed under the Cityscapes → Foggy Cityscapes setting unless otherwise specified. The KITTI → Cityscapes scenario was used exclusively in the SAM-based segmentation supervision experiments.
    \subsubsection{Effects of mask ratio}
\begin{figure} 
    \centering
    \includegraphics[width=\linewidth]{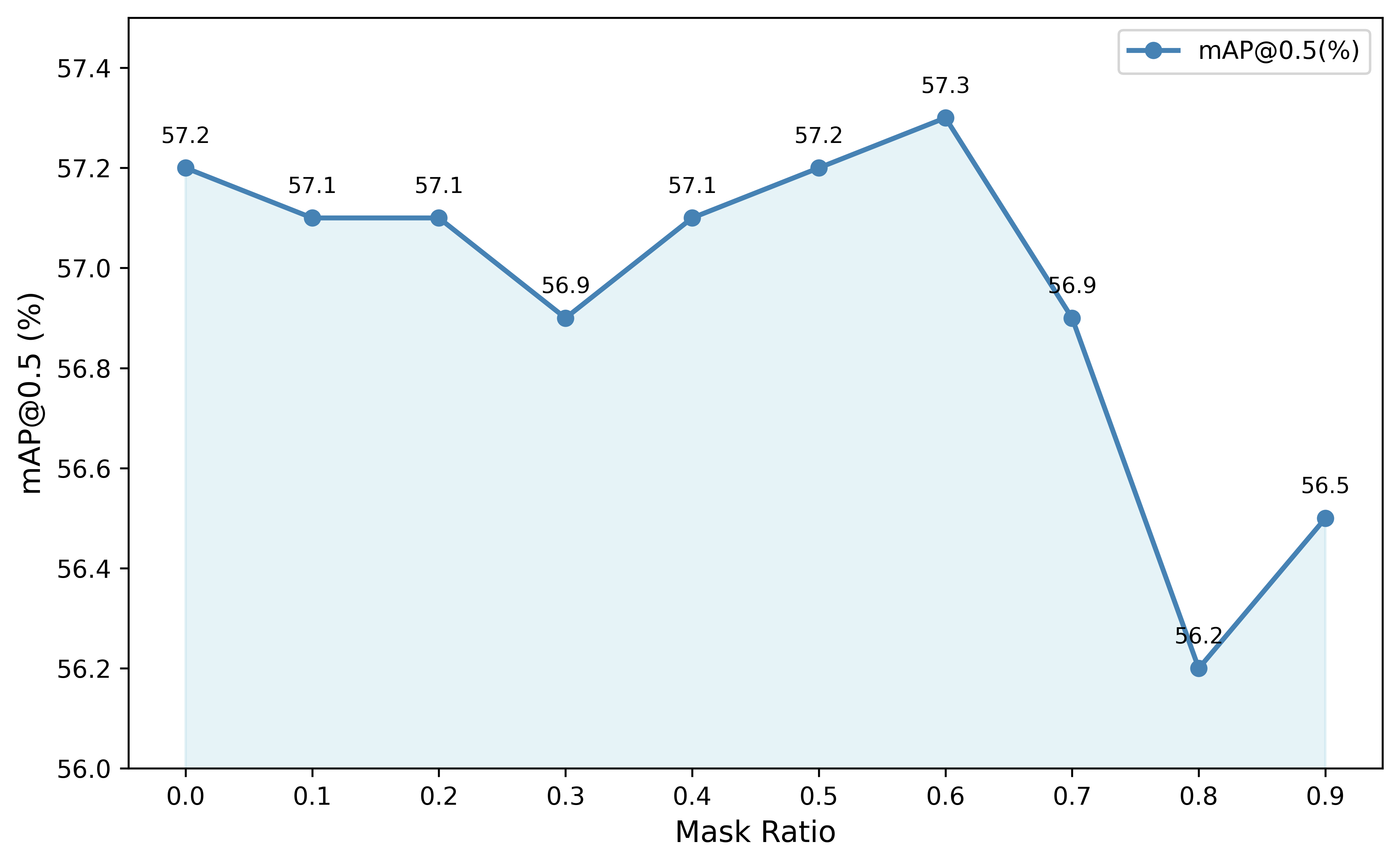}
    \caption{Effect of different mask ratios on model training. Moderate masking improves model performance, while excessive masking considerably declines the performance.}
    \label{picture/7.png}
\end{figure}

    Experiments were conducted under varying mask ratio ranging from 0 to 0.9 to assess the impact of the mask ratio applied during random occlusion. All other settings remain unchanged. As the mask ratio gradually increased from zero, the performance of the model remained stable or slightly improved under moderate masking ($mask\ ratio\in[0.1, 0.6]$) (Fig. \ref{picture/7.png}). This indicated that the masking mechanism, up to a certain extent, could be an optimizer, which would enhance the robustness of the model when local information would be missing. However, the performance considerably declined when the mask ratio continued to increase ($\geq 0.7$), particularly when the mask ratio was 0.8, where mAP dropped to its lowest value. This suggests that excessive masking could disrupt the semantic integrity of the input image, preventing the student model from effectively aligning with the predictions of the teacher model. Based on comprehensive experimental results, we selected a mask ratio of 0.6 as the default configuration to maintain the robustness of the model and the integrity of semantic information. Under this setting, the detection performance of the student model on the target domain was the most stable, with the capability to effectively improve adaptability to local occlusion and noise while retaining sufficient semantic features.

    \subsubsection{Hyperparameter sensitivity: loss weight configuration}
\begin{figure}
    \centering
    \includegraphics[width=\linewidth]{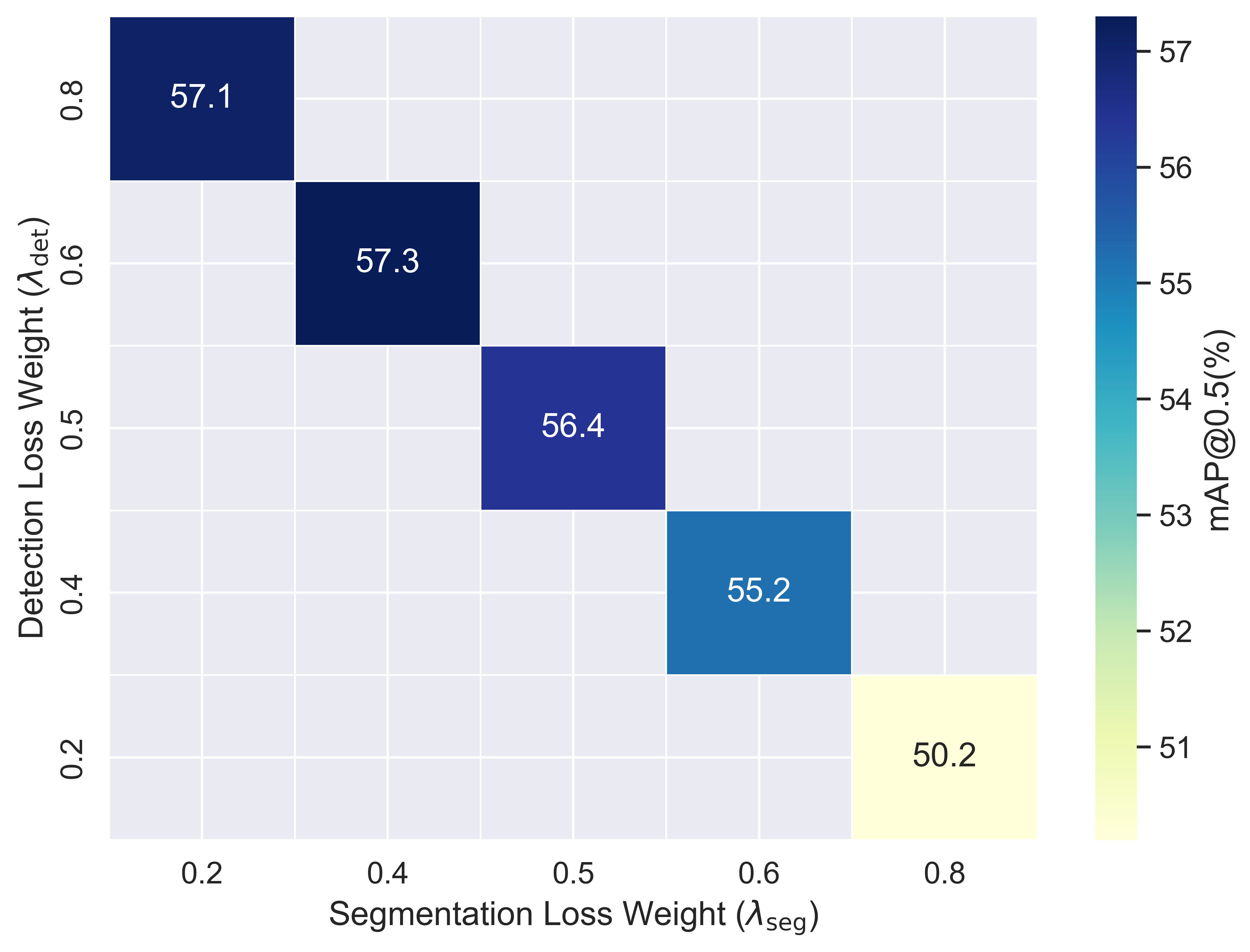}
    \caption{Effect of $\lambda_{\mathrm{det}}$ and $\lambda_{\mathrm{seg}}$  combinations on the mAP of the model in the target domain. When the weights are set to $\lambda_{\mathrm{det}}$ = 0.6 and $\lambda_{\mathrm{seg}}$ = 0.4, the model achieves the best performance (57.3\%).}
    \label{picture/8.png}
\end{figure}

    The joint optimization objective included weighted contributions from detection and segmentation losses. We determined the optimal weights by varying the detection loss weight $\lambda_{\mathrm{det}}$  and segmentation loss weight $\lambda_{\mathrm{seg}}$  while keeping other parameters constant. The highest detection accuracy (mAP = 57.3\%) was achieved when $\lambda_{\mathrm{det}}$ = 0.6 and $\lambda_{\mathrm{seg}}$ = 0.4 (Fig. \ref{picture/8.png}). This configuration emphasized detection while leveraging meaningful guidance from the auxiliary segmentation task. When the segmentation weight was reduced (e.g., $\lambda_{\mathrm{seg}}$ = 0.2), performance improvement was limited, indicating insufficient contribution from the auxiliary task. Conversely, a high segmentation weight (e.g., $\lambda_{\mathrm{seg}}$ = 0.8) considerably degraded the performance (mAP = 50.2\%), likely owing to an overemphasis on pixel-level structure at the expense of object-level localization. Thus, setting $\lambda_{\mathrm{det}}$ = 0.6 and $\lambda_{\mathrm{seg}}$ = 0.4 effectively balanced tasks and was selected for subsequent experiments.

    \subsubsection{Validity of the proposed module}
    We assessed the individual and combined contributions of the proposed components by conducting a series of component-wise ablation studies using SSDA-YOLO as the baseline framework. The results are summarized in Table \ref{tab:4}.\par
    Initially, a source only model, YOLOv5-L, trained on the source domain under full supervision and tested directly on the target domain was evaluated. This model achieved a mAP of only 35.9\%, which indicated its poor generalization performance under considerable domain shift. Building upon this, the SSDA-YOLO framework was adopted as a baseline. The introduction of a teacher-student consistency mechanism to mitigate domain discrepancies highly improved the mAP to 55.9\%, confirming the effectiveness of semi-supervised consistency–based adaptation.\par

\begin{table*}[ht]
\centering
\caption{ Impact of different modules on detection performance. Starting from Source Only, adding Baseline (SSDA-YOLO), Masking (masking mechanism), and DenseRFB modules (auxiliary segmentation head) progressively improves results.}
\label{tab:4}
\small
\begin{tabular}{cccc|cc}
\hline
\textbf{Source Only} & \textbf{Baseline} & \textbf{Masking} & \textbf{DenseRFB} & \textbf{mAP@0.5(\%)} & \textbf{Gain(\%)} \\
\hline
\checkmark &  &  &  & 35.9 & - \\
\checkmark & \checkmark &  &  & 55.9 & +20.0 \\
\checkmark & \checkmark & \checkmark &  & 56.8 & +20.9 \\
\checkmark & \checkmark &  & \checkmark & 57.2 & +21.3 \\
\checkmark & \checkmark & \checkmark & \checkmark & \textbf{57.3} & \textbf{+21.4} \\
\hline
\end{tabular}
\end{table*}

    To further enhance the robustness of the model, a masking mechanism was incorporated into the baseline, along with a corresponding mask-consistency loss. This encouraged the student model to attend to structurally complex regions of the image, which improved its performance to 56.8\%. In parallel, a semantic segmentation head based on the proposed DenseRFB was introduced as an auxiliary task. Without incorporating the masking mechanism, this module alone improved the mAP to 57.2\%, suggesting that the pixel-level structural supervision can serve as a strong prior under domain shift. Finally, when both the components—the mask-consistency loss and segmentation supervision—were integrated into the SSDA-YOLO framework, the model achieved the highest mAP of 57.3\%. This represented a +21.4\% improvement over the source only model and outperformed the standard SSDA-YOLO baseline.\par
    These results conclusively demonstrated that the masking mechanism and auxiliary segmentation head independently contributed to cross-domain performance gains, and that their combination additively enhanced the generalization capability of the model.
    
    \subsubsection{Grad-CAM visualization}
\begin{figure*}[ht]
    	\centering
    	\includegraphics[width=0.9\textwidth]{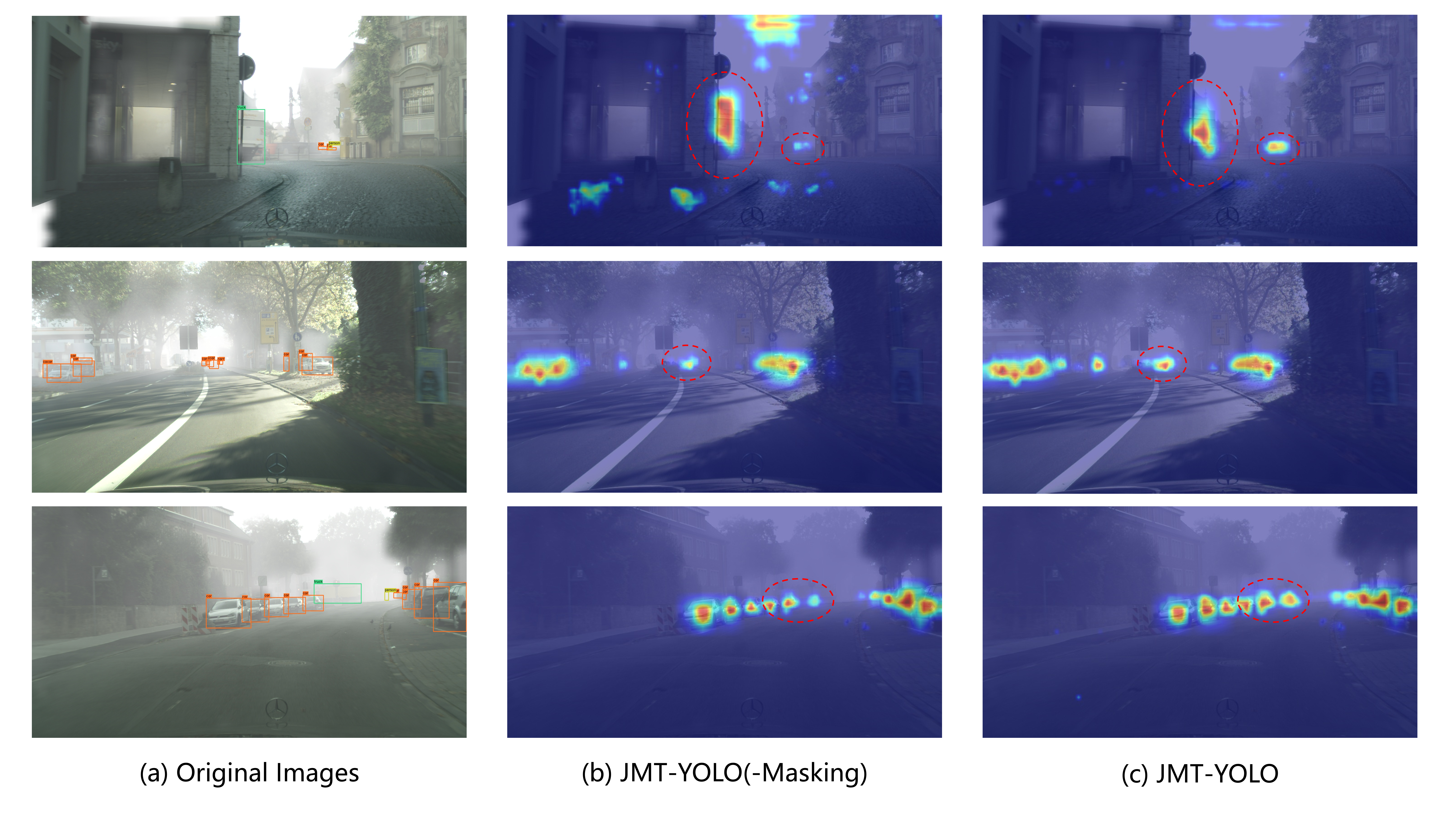}
    	\caption{Grad-CAM visualizations on Foggy Cityscapes. With masking, the activations become more compact and aligned with object regions, while the unmasked model shows weaker responses.}
    	\label{picture/9.png}
    \end{figure*}

    The effectiveness of the proposed masking mechanism was further examined through Grad-CAM visualizations on Foggy Cityscapes. As shown in Fig. \ref{picture/9.png}, masking noticeably improved instance-level alignment under fog degradation. In the first row, heavy fog caused the truck to visually merge with the background building; without masking, the model produced expanded and ambiguous activations that covered both regions and yielded weak responses on the distant car due to fog attenuation. With masking, the activations became compact and well aligned with the actual object extents, clearly separating the truck from the background and enhancing the response to the far car. In the second and third rows, the unmasked model exhibited weaker activations under fog, whereas the masked model generated more complete and object-focused heatmaps. These results indicated that masking encouraged global contextual reasoning and improved semantic consistency across instances in the target domain.
    
    \subsubsection{Prompting strategies for SAM pseudo masks}
    \label{sec:sam}
    To mitigate the scarcity of pixel-level annotations, SAM-generated pseudo masks were used to provide scalable supervision for the auxiliary segmentation branch. In addition, the effectiveness of this supervision was evaluated when high-quality GT box prompts were unavailable. Specifically, SAM pseudo masks were generated on the source-domain KITTI data under different prompting strategies, and the proposed method was evaluated under the KITTI → Cityscapes (car) setting.\par
    
    As shown in Table \ref{tab:5}, the baseline achieved 59.6\% mAP@0.5 and 29.8\% mAP@[.5:.95]. Supervising the segmentation branch with partial masks (200 GT masks) improved the performance to 61.0\%/30.6\%, indicating that even limited pixel-level supervision introduced useful structural constraints for segmentation-guided consistency learning. Replacing partial masks with full-dataset SAM pseudo masks prompted by GT boxes yielded the best results (62.0\%/32.4\%), suggesting that denser and better-aligned pseudo masks provided more reliable shape cues and enhanced the regularization effect of the auxiliary segmentation branch for detection.\par
    
\begin{figure*}[t]
	\centering
	\includegraphics[width=\linewidth]{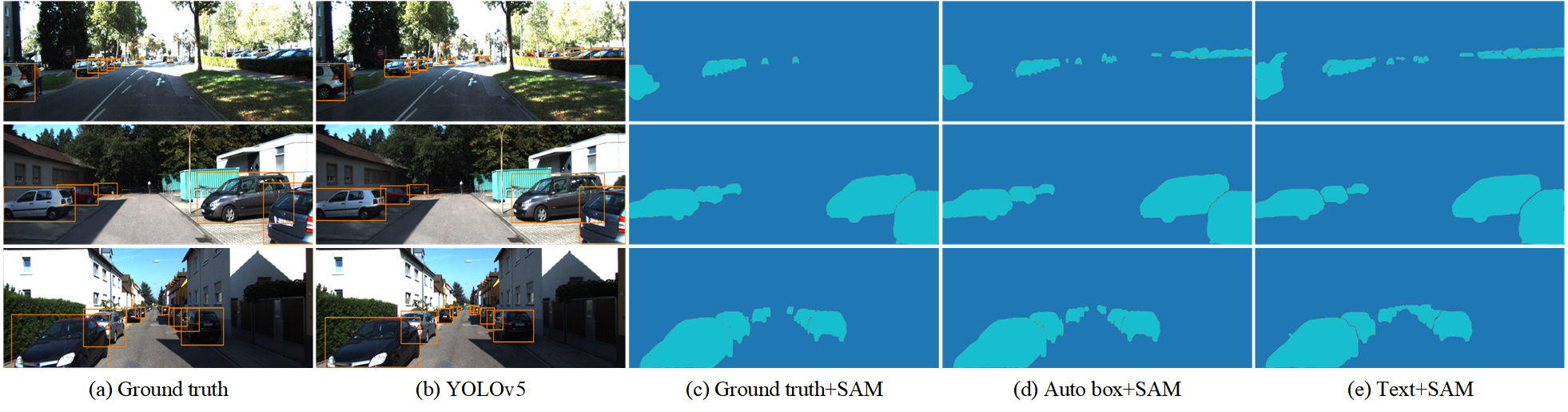}
	\caption{Qualitative comparison of SAM pseudo masks generated with different prompting strategies on KITTI (car). GT box prompts generally produce more complete and boundary-aligned masks. Auto box prompts remain effective but can introduce slight boundary noise. Text prompts are less spatially constrained, leading to occasional missing or fragmented masks in complex scenes.}
	\label{picture/sam_prompt_vis.png}
\end{figure*}

    The gains were generally preserved under weaker prompts. Specifically, we used a COCO-pretrained YOLOv5-L detector to generate predicted boxes as SAM prompts (auto box+SAM), and directly adopted the text prompt “car with wheels and windows” for box-free SAM prompting (text+SAM). Relative to the baseline, auto box+SAM improved mAP@0.5/mAP@[.5:.95] by +2.1/+2.5(61.7\%/32.3\%), while text+SAM improved it by +1.3/+1.3 (60.9\%/31.1\%). As illustrated in Fig.~\ref{picture/sam_prompt_vis.png}, the predicted-box prompts provided sufficient localization cues, enabling SAM to recover masks for hard cases (e.g., distant or partially occluded cars) and thereby enriching the structural supervision. Meanwhile, the descriptive text prompt still enabled SAM to segment meaningful car regions even without explicit box constraints (Fig.~\ref{picture/sam_prompt_vis.png}). Overall, these results suggested that SAM-based pseudo masks could serve as an effective source of segmentation supervision for the auxiliary branch when high-quality GT box prompts were unavailable.\par

\begin{table}[t]
	\centering
	\caption{Comparison of SAM prompting strategies for pseudo-mask supervision on KITTI → Cityscapes (car), including GT box prompts, auto-predicted box prompts, and a box-free text prompt.}
	\label{tab:5}
	\small
	\setlength{\tabcolsep}{4pt}
	\begin{tabular}{lcc}
		\toprule
		\textbf{Status} & \textbf{mAP@0.5(\%)} & \textbf{mAP@[.5:.95](\%)} \\
		\midrule
		Baseline            & 59.6 & 29.8 \\
		Ours (200 GT masks) & 61.0 & 30.6 \\
		Ours (GT+SAM)       & \textbf{62.0} & \textbf{32.4} \\
		Ours (auto box+SAM) & 61.7 & 32.3 \\
		Ours (text+SAM)     & 60.9 & 31.1 \\
		\bottomrule
	\end{tabular}
\end{table}

	\subsubsection{Robustness to noisy box prompts}
	\label{sec:sam_noisy}
	Robustness to imperfect box prompts was further evaluated by perturbing the GT box centers with different center-shift magnitudes (noisy$\pm r\%$). As reported in Table~\ref{tab:6}, performance degraded as the perturbation increased. However, under mild to moderate shifts (e.g., noisy$\pm$10\% and noisy$\pm$30\%), the proposed method still maintained gains over the baseline, indicating that SAM tolerated a certain level of prompt misalignment and continued to provide usable pseudo-mask supervision.\par
	
\begin{figure*}[t]
	\centering
	\includegraphics[width=\linewidth]{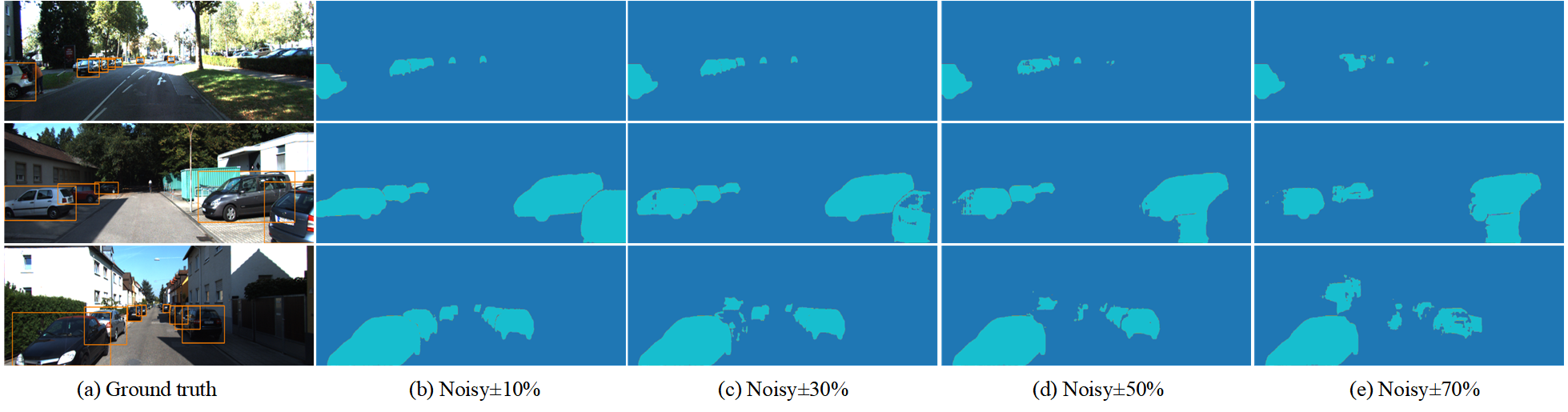}
	\caption{Effect of box-prompt center-shift noise on SAM pseudo masks (noisy$\pm r\%$) on KITTI (car). Mild shifts preserve sufficient overlap and yield usable masks, whereas large shifts drift toward background or adjacent regions, resulting in fragmented or incorrect masks.}
	\label{picture/sam_noisy_vis.png}
\end{figure*}

	As shown in Fig.\ref{picture/sam_noisy_vis.png}, when the perturbation was small, the prompt boxes still overlapped sufficiently with the targets, and SAM typically produced usable masks that roughly matched the object extent. When the perturbation became large (e.g., $\ge$50\%), the prompts drifted toward background or adjacent regions, resulting in fragmented or incorrect masks, which reduced the informativeness of supervision and could even become misleading. The degradation was more pronounced on mAP@[.5:.95](Table~\ref{tab:6}), since higher IoU thresholds were more sensitive to boundary accuracy and coverage completeness. Overall, these results indicated that SAM-based pseudo-mask supervision remained effective under mildly noisy prompts, while severe misalignment substantially diminished its utility.

\begin{table}[t]
	\centering
	\caption{Effect of box-prompt noise on SAM-based pseudo-mask supervision on KITTI → Cityscapes (car), where noisy$\pm r\%$ denotes $\pm r\%$ center-shift perturbation of GT box prompts.}
	\label{tab:6}
	\small
	\setlength{\tabcolsep}{4pt}
	\begin{tabular}{lcc}
		\toprule
		\textbf{Status} & \textbf{mAP@0.5(\%)} & \textbf{mAP@[.5:.95](\%)} \\
		\midrule
		Baseline          & 59.6 & 29.8 \\
		Ours (GT+SAM)     & \textbf{62.0} & \textbf{32.4} \\
		Ours (noisy$\pm$10\%) & 61.0 & 31.8 \\
		Ours (noisy$\pm$30\%) & 60.7 & 30.9 \\
		Ours (noisy$\pm$50\%) & 60.4 & 29.7 \\
		Ours (noisy$\pm$70\%) & 58.6 & 28.6 \\
		\bottomrule
	\end{tabular}
\end{table}

     \subsubsection{t-SNE visualization}
\begin{figure*}[ht]
    \centering
    \includegraphics[width=0.9\textwidth]{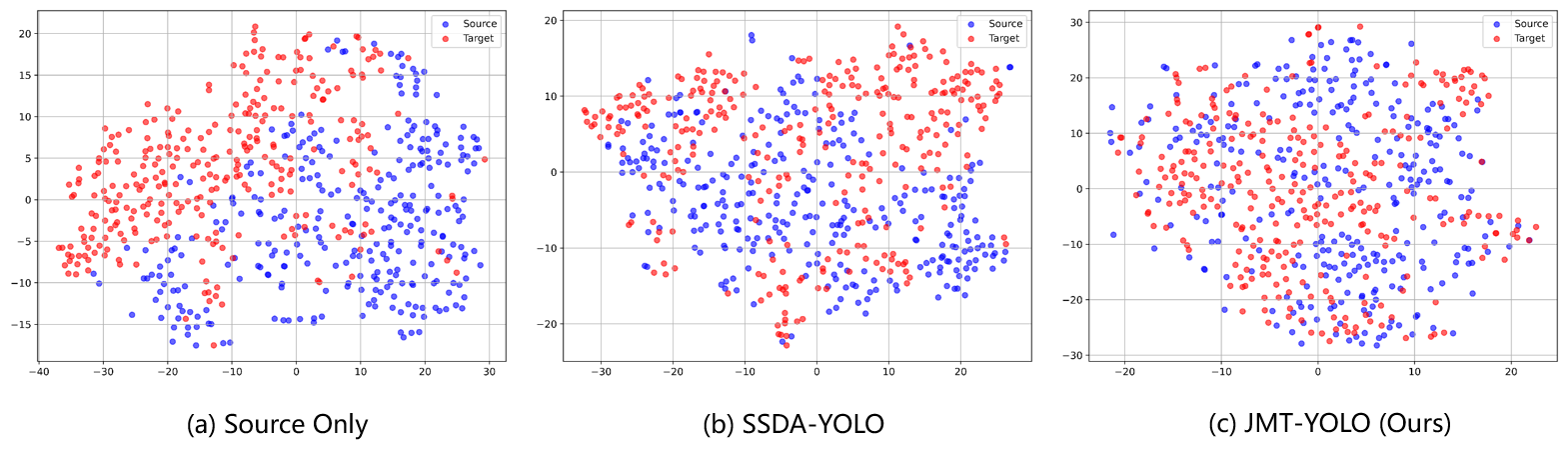}
    \caption{t-SNE visualization comparison of intermediate feature distributions. (a) Source Only model, source domain (blue), and target domain (red) features are clearly separated, (b) SSDA-YOLO model, source and target domain features begin to merge but differences still exist, (c) our proposed method JMT-YOLO, source and target domain feature distributions highly overlapped, indicating stronger domain alignment capability.}
    \label{picture/10.png}
\end{figure*}

     To assess feature alignment in the latent space, t-SNE was used to visualize the feature distributions of source and target samples (Fig.\ref{picture/10.png}). In Fig. \ref{picture/10.png}(a), for the source only model, the source domain features (blue) and target-domain features (red) showed a clear separation in distribution, indicating the model’s failure in effectively generalizing the target-domain. Fig. \ref{picture/10.png}(b) illustrates the feature distribution of the SSDA-YOLO model, where source and target-domain samples are noticeably mixed, suggesting that introducing the teacher-student consistency supervision mechanism effectively alleviated interdomain differences. However, some degree of distribution shift existed. By contrast, Fig. \ref{picture/10.png}(c) presents the feature distribution of JMT-YOLO. The source and target-domain features were highly fused in the embedding space, with blurred boundaries and a higher degree of intermingling. This indicated that our designed auxiliary segmentation task and masking mechanism played a key role in enhancing the representation capability of target-domain features and reducing domain differences, further improving the generalization ability of the model.
     
	\section{Conclusions}
	\label{Conclusions}
    This study addressed the crucial challenge of domain shift in DAOD by proposing JMT-YOLO, a novel joint multi-task learning framework. Developed upon the MT framework, the proposed method integrated two key components: an auxiliary segmentation task and a masking mechanism. These components were designed to enhance structural feature representation and instance-level alignment between the source and target domains. Experimental results on multiple DAOD tasks conclusively demonstrated that the proposed method exhibited considerable advantages in typical complex scenarios, fully validating its effectiveness and generalization ability. We additionally evaluated SAM-generated pseudo masks to complement scarce segmentation supervision, and observed that the gains are largely maintained under weaker or box-free prompts. Although our proposed method demonstrated noticeable performance improvements, certain limitations exist. In particular, the reliability of pseudo-labels could considerably affect the final performance. Future studies would focus on investigating more robust pseudo-label generation techniques to mitigate label noise and further improve model stability and robustness. 
    
	\section*{CRediT authorship contribution statement}
    \textbf{Zhenhua Yu}: Conceptualization, Supervision, Writing – original draft, Funding acquisition. \textbf{Jiahao Zhang}: Methodology, Software, Validation, Writing – original draft. \textbf{Wenchao Zhang}: Conceptualization, Methodology, Writing – review \& editing. \textbf{Ye Ou}: Formal analysis, Writing – review \& editing. \textbf{Hengxiang Zhao}: Data curation

    \section*{Declaration of competing interest}
    The authors declare that they have no known competing financial interests or personal relationships that could have appeared to influence the work reported in this paper.

    \section*{Acknowledgements}
    This work was supported in part by the Natural Science foundation of China under Grant 62273272, Grant 62303375, and Grant 61873277; in part by the Key Research and Development Program of Shaanxi Province under Grant 2024CY2-GJHX-49 and Grant 2024CY2-GJHX-43; in part by the Youth Innovation Team of Shaanxi Universities; and in part by the Key Scientific Research Program of Education Department of Shaanxi Province under Grant 24JR111.

    \section*{Data availability}
    Data will be made available on request.

	
	
	
%
	
	
	

	
%
%
	
\bibliographystyle{model3-num-names} 
\bibliography{reference1}

\end{document}